\documentclass[unnumsec,webpdf,contemporary,large]{oup-authoring-template}%
\usepackage{graphicx,url}
\usepackage{xspace}
\usepackage{xcolor} %highlight text
\usepackage{longtable}
\usepackage{multirow}

\theoremstyle{thmstyleone}%
\theoremstyle{thmstyletwo}%
\theoremstyle{thmstylethree}%

\newcommand{\ie}{\emph{i.e.,}\xspace}
\newcommand{\eg}{\emph{e.g.,}\xspace}
\newcommand{\etal}{\emph{et al.}\xspace}

\newcolumntype{L}[1]{>{\raggedright\arraybackslash}p{#1}}
 
\newcolumntype{C}[1]{>{\centering\arraybackslash}p{#1}}
 
\newcolumntype{R}[1]{>{\raggedleft\arraybackslash}p{#1}}

\begin{document}

\journaltitle{Preprint}
%\DOI{DOI HERE}
\copyrightyear{2021}
\pubyear{2021}
%\access{Advance Access Publication Date: Day Month Year}
\appnotes{Reviews}

\firstpage{1}

%\subtitle{Subject Section}
% %Before uploading your manuscript, please ensure that you have included the following:
% A short abstract, outlining the aims and subject matter.
% A biographical note of approximately 30 words.
% Three to five key points summarising the key messages of your article.
% Up to six keywords for indexing purposes.

\title[Machine learning for prediction of cancer drivers]{Machine learning methods for prediction of cancer driver genes: a survey paper\footnote{Preprint. This work has been submitted to Oxford University Press for possible publication.}}

\author[1,2]{Renan Andrades}
\author[1,2,$\ast$]{ Mariana Recamonde-Mendoza}
%\author[3]{Third Author}
%\author[3]{Fourth Author}
%\author[4]{Fifth Author}

\authormark{Andrades and Recamonde-Mendoza}

\address[1]{\orgdiv{Institute of Informatics}, \orgname{Universidade Federal do Rio Grande do Sul}, \orgaddress{\state{Porto Alegre/RS}, \country{Brazil}}}
\address[2]{\orgdiv{Bioinformatics Core}, \orgname{Hospital de Clínicas de Porto Alegre}, \orgaddress{\state{Porto Alegre/RS}, \country{Brazil}}}
%\address[3]{\orgdiv{Department}, \orgname{Organization}, \orgaddress{\street{Street}, \postcode{Postcode}, \state{State}, \country{Country}}}
%\address[4]{\orgdiv{Department}, \orgname{Organization}, \orgaddress{\street{Street}, \postcode{Postcode}, \state{State}, \country{Country}}}

\corresp[$\ast$]{Corresponding author. \href{mrmendoza@inf.ufrgs.br}{mrmendoza@inf.ufrgs.br}}

% \received{Date}{0}{Year}
% \revised{Date}{0}{Year}
% \accepted{Date}{0}{Year}

%\editor{Associate Editor: Name}

%\abstract{
%\textbf{Motivation:} .\\
%\textbf{Results:} .\\
%\textbf{Availability:} .\\
%\textbf{Contact:} \href{name@bio.com}{name@bio.com}\\ 

%\textbf{Supplementary information:} Supplementary data are available at \textit{Briefings in Bioinformatics}
%online.}

\abstract{Identifying the genes and mutations that drive the emergence of tumors is a major step to improve understanding of cancer and identify new directions for disease diagnosis and treatment. Despite the large volume of genomics data, the precise detection of driver mutations and their carrying genes, known as cancer driver genes, from the millions of possible somatic mutations remains a challenge. Computational methods play an increasingly important role in identifying genomic patterns associated with cancer drivers and developing models to predict driver events. Machine learning (ML) has been the engine behind many of these efforts and provides excellent opportunities for tackling remaining gaps in the field. Thus, this survey aims to perform a comprehensive analysis of ML-based computational approaches to identify cancer driver mutations and genes, providing an integrated, panoramic view of the broad data and algorithmic landscape within this scientific problem. We discuss how the interactions among data types and ML algorithms have been explored in previous solutions and outline current analytical limitations that deserve further attention from the scientific community. We hope that by helping readers become more familiar with significant developments in the field brought by ML, we may inspire new researchers to address open problems and advance our knowledge towards cancer driver discovery.}

% \boxedtext{
% \begin{itemize}
% \item Key boxed text here.
% \item Key boxed text here.
% \item Key boxed text here.
% \end{itemize}}

\maketitle

\section{Introduction}

Cancer is among the leading causes of morbidity and mortality worldwide irrespective of the level of human development, accounting for nearly 10 million deaths in 2020 \cite{CAtoday}. According to GLOBOCAN 2020 estimates, 28.4 million new cancer cases are projected to occur in 2040 globally, representing a 47\% increase from the number of cases in 2020  \cite{bray2018global}.
One of the most promising directions in reducing cancer mortality is the early detection and treatment of cancer lesions, which demands a better understanding of the molecular mechanisms underlying tumor emergence and progression \cite{loomans2019cancer, beane2017genomic}.

Although several factors may contribute to carcinogenesis, such as epigenetic modifications and tumor environment, cancer is primarily a result of genetic alterations \cite{anandakrishnan2019estimating}. The disease is characterized by abnormal and uncontrolled cell growth caused by somatic mutations, which encompass several distinct classes of changes in DNA sequence, from single nucleotide variants (SNVs) to small insertions and deletions (indels) and larger structural variations \cite{martinez2020compendium, stratton2009cancer}. The somatic mutations that confer a selective growth advantage to a tumor cell, assuming causal roles in the initiation or progression of cancer, are called ``driver” mutations \cite{stratton2009cancer}. These mutations reside in a subset of genes known as \textit{cancer driver genes} (CDGs). However, not all somatic mutations are causally implicated in carcinogenesis. In fact, the vast majority of these alterations have no impact on the %no direct or indirect effect for 
initiation and evolution of cancer, being denominated ``passenger" mutations \cite{stratton2009cancer, vogelstein2013cancer}. One of the main goals in cancer research is the identification of all genes carrying mutations able to drive carcinogenesis across different tumor types, a knowledge that has a major impact from disease diagnosis to personalized therapies. This goal, however, involves the challenging task of distinguishing driver mutations from passenger mutations \cite{martinez2020compendium}. 

Many computational methods have been developed in the last years to pinpoint mutations and genes with causal relation with cancer from genomics data, allowing significant advances in the comprehensive identification of CDGs \cite{martinez2020compendium, bailey2018comprehensive}.  
Among the several possible strategies, machine learning (ML) methods are receiving increasing attention due to their outstanding performance on several prediction tasks in bioinformatics \cite{greener2021guide}. 
However, distinct methods employ different molecular features and computational algorithms to build predictive models. As pointed by previous works \cite{tamborero2013comprehensive, chen2020comprehensive}, these differences in the conception, implementation, and prediction target %(\eg cancer-specific or not) 
influence methods' findings, such that usually only a partial overlap is observed among their outputs. % of distinct approaches for identifying driver mutations and CDG. %This requires a posterior curation of results by an expert or by wet-lab experiments to reduce the false-positive rate. 
Thus, it is important to know the plethora of options available and their particularities, including their respective strengths and weaknesses, to choose the most suitable method. % for the systematic detection of candidate drivers.

Previous works have dedicated efforts to summarize and organize literature regarding computational approaches for prediction of CDGs and driver mutations, with different focuses \cite{zhang2014identifying,chen2015deciphering,cheng2016advances,dimitrakopoulos2017computational,chen2020comprehensive,Rogers+2021}.
Although some ML-based methods were covered by some of these works, they were mostly developed for a more general problem of distinguishing disease-related SNVs from common polymorphisms \cite{zhang2014identifying,chen2015deciphering}. Other surveys have categorized methods according to their major feature types  \cite{cheng2016advances} or prediction strategy \cite{dimitrakopoulos2017computational}, without emphasis on ML. Recent works also focused on comparing the predictive performance of distinct computational methods for cancer drivers prediction \cite{chen2020comprehensive, pham2021computational}. Finally, Rogers \etal \cite{Rogers+2021} reviewed generic ML-based tools to predict the pathogenic impact of human genome variants, further concentrating their discussion on a specific set of tools to predict cancer drivers. %variants that may act as enablers of unregulated cell proliferation. 
Nonetheless, the theoretical background and methodological details of ML were not discussed by the authors.

All these previous works, jointly, have covered a large body of the literature regarding computational methods for cancer driver discovery and have been crucial to elucidate the particular niche targeted by each solution, as well as their potential in solving the task. %Their efforts have been crucial to elucidate important methodological details and the particular niche targeted by each solution, guiding readers towards a better exploration of these methods in their studies. 
We emphasize, however, that aspects entailed in the development of ML models for CDGs prediction were not the focus of previous discussions. %none of these efforts have focused on reviewing ML-based approaches. Thus, despite the large variety of methods reviewed, aspects entailed in the development of ML models for CDG prediction were not the focus of previous discussions. %Given the increasing popularity of ML algorithms in this domain motivated by their promising performance, a better understanding of ML methods is an essential step towards their effective exploration in this task. %the task of identifying cancer drivers.%, and equally important, for further methodological advance of the field.
Therefore, the present survey aims to provide a comprehensive analysis of ML-based computational approaches to identify driver mutations and CDG, constructing an integrated, panoramic view of data and ML techniques within this domain. In particular, we focus on summarizing the main trends in relation to the types of algorithms and data (\ie model features) employed, the specific target of prediction, strategies for models evaluation, and their potential performance for the task of interest. We conclude by pointing open issues and proposing future research directions in this field.%We also aim at analyzing the advantages and disadvantages of current approaches, identifying specific gaps in the related literature that deserve special attention in future works, and pointing promising research directions.

\begin{table*}[!th]
\caption{Main characteristics of selected papers. \label{tab:selectedPapers}}
\centering
\resizebox{\textwidth}{!}{%
\begin{tabular}{llclcll}
\hline
Reference                   & Data categories$^{1}$      & \multicolumn{1}{l}{Data integration} & Algorithms$^{2}$       & \multicolumn{1}{l}{Feature selection} & Performance metrics$^{3}$             & Validation strategy          \\ \hline
Carter \etal (2009) \cite{Carter2009}        & GV;FI          & Y                                    & Tb              & Y                                     & ROC; Acc                       & k-fold CV                       \\
Capriotti and Altman (2011) \cite{Capriotti2011} & GV;FI;Ob       & Y                                    & SVM             & N                                     & ROC; Acc                        & k-fold CV                       \\
Fu \etal (2012) \cite{Fu2012}            & FG             & N                                    & So              & N                                     &                                 &                                 \\
Tan \etal (2012) \cite{Tan2012}           & GV;FI;Ob             & Y                                    & SVM             & Y                                     & Acc                             & k-fold CV                       \\
Davoli \etal (2013) \cite{Davoli2013}        & GV;FI          & Y                                    & Rb              & Y                                     & Acc                             & k-fold CV                       \\
Mao \etal (2013) \cite{Mao2013}           & GV;FI          & Y                                    & SVM             & Y                                     & ROC                             & k-fold CV                       \\
Manolakos \etal (2014) \cite{Manolakos2014}     & FG;Ob             & Y                                    & UL              & N                                     &                                 & Holdout (70\%/30\%) + k-fold CV \\
Schroeder \etal (2014) \cite{Schroeder2014}     & GV;FI          & Y                                    & Rb;Tb;Pm        & N                                     & Acc                             & k-fold CV                       \\
U \etal (2014) \cite{U2014}             & FI             & N                                    & SVM;Tb;NN;Pm;So & Y                                     & Acc; F1; Pre; Rec     & k-fold CV                       \\
Anoosha \etal (2015) \cite{Anoosha2015}       & FI          & N                                    & SVM             & Y                                     & Acc                             & k-fold CV                       \\
Gnad \etal (2015) \cite{Gnad2015}          & GV;FI;FG       & Y                                    & Rb              & Y                                     & Acc                             & k-fold CV                       \\
Park \etal (2015) \cite{Park2015}          & GV;FG;Nb       & Y                                    & Rb              & Y                                     & Acc                             &                                 \\
Soliman \etal (2015) \cite{Soliman2015}       & GV;FI             & Y                                    & SVM;Rb          & Y                                     & Acc; F1; MCC                         & k-fold CV                       \\
Dong \etal (2016) \cite{Dong2016}          & GV;FI;Ob       & Y                                    & SVM;Rb          & Y                                     & ROC                             & k-fold CV                       \\
Li \etal (2016) \cite{LI2016}            & GV             & N                                    & EA              & N                                     &                                 &                                 \\
Tokheim \etal (2016) \cite{Tokheim2016}       & GV;FI;FG;Nb       & Y                                    & Tb              & N                                     &                             &                                 \\
Park \etal (2017) \cite{Park2017}          & GV;FG          & Y                                    & Rb              & Y                                     & Acc                             &                                 \\
Tavanaei \etal (2017) \cite{Tavanaei2017}      & FI             & N                                    & DL              & N                                     & ROC; Acc; Pre; Rec     & Holdout (85\%/15\%)             \\
Agajanian \etal (2018) \cite{Agajanian2018}     & FI          & N                                    & Rb;Tb           & Y                                     & Acc; F1; Pre; Rec      & Holdout (80\%/20\%) + k-fold CV \\
Celli \etal (2018) \cite{Celli2018}         & FG             & N                                    & Tb              & N                                     & F1                              & Holdout (70\%/30\%)             \\
Guan \etal (2018) \cite{Guan2018}          & GV;FG          & Y                                    & SVM             & Y                                     &                                 & k-fold CV                       \\
Lu \etal (2018) \cite{Lu2018}            & GV;FG          & Y                                    & UL              & N                                     & Acc; F1; Rec                 & k-fold CV                       \\
Wang \etal (2018) \cite{Wang2018}          & GV;FI;FG       & Y                                    & Pm              & N                                     & ROC                             & k-fold CV                       \\
Zhou \etal (2018) \cite{Zhou2018}          & FI;Ob;Nb    & Y                                    & Tb              & Y                                     & F1; Rec; MCC                     & k-fold CV                       \\
Agajanian \etal (2019) \cite{Agajanian2019}     & GV;FI          & Y                                    & Tb;DL           & Y                                     & ROC; Acc; F1                    & Holdout (80\%/20\%) + k-fold CV \\
Althubaiti \etal (2019) \cite{Althubaiti2019}    & Ob             & N                                    & NN              & N                                     & ROC; F1                         & k-fold CV                       \\
Collier \etal (2019) \cite{Collier2019}       & GV;FI;Nb          & Y                                    & SVM             & N                                     & ROC                             & k-fold CV                       \\
Han \etal (2019) \cite{Han2019}           & GV             & N                                    & So              & N                                     &                                 &                                 \\
Jiang \etal (2019) \cite{Jiang2019}         & GV;FG          & Y                                    & Tb              & N                                     &                            &                                 \\
Luo \etal (2019) \cite{Luo2019}           & GV;FG          & Y                                    & DL              & N                                     & ROC                             &                                 \\
Nicora \etal (2019) \cite{Nicora2019}       & GV;FI          & Y                                    & SVM;Rb;Tb       & N                                     & Acc; Rec                     & Holdout (70\%/30\%) + k-fold CV \\
Schulte-Sasse \etal (2019) \cite{Schulte-Sasse2019} & GV;FG;Nb       & Y                                    & DL              & N                                     & ROC; PRC; MCC                   & k-fold CV                       \\
Xi \etal (2019) \cite{Xi2019}            & GV             & N                                    & UL              & N                                     & Pre; Rec               & k-fold CV                       \\
Zhu \etal (2019) \cite{Zhu2019}           & FI             & N                                    & -         & N                                     &                                 &                                 \\
Chandrashekar \etal (2020) \cite{Chandrashekar2020} & GV;FI             & Y                                    & Tb              & N                                     & Acc                             & k-fold CV                       \\
Colaprico \etal (2020) \cite{Colaprico2020}     & GV;FI;FG;Ob;Nb       & Y                                    & Tb              & N                                     &                                 &                                 \\
Cutigi \etal (2020) \cite{Cutigi2020}        & GV;Nb          & Y                                    & SVM;Tb          & N                                     & ROC; Acc; F1; Pre; Rec & k-fold CV                       \\
Gumpinger \etal (2020) \cite{Gumpinger2020}     & FI;Nb             & Y                                    & SVM;Rb;Tb       & N                                     & F1; Pre; Rec; PRC  & k-fold CV                       \\
Lyu \etal (2020) \cite{Lyu2020}           & GV;FI;FG;Ob    & Y                                    & SVM;Rb;Tb       & Y                                     & Acc; Pre; Rec; PRC & k-fold CV                       \\
Wang \etal (2020) \cite{Wang2020}           & FI    & N                                    & SVM;Tb;NN;So       & N                                     & ROC; Acc; F1; Pre; Rec & k-fold CV                       \\
Nulsen \etal (2021) \cite{Nulsen2021}        & GV;FI;FG;Ob;Nb & Y                                    & SVM             & Y                                     & ROC                             & k-fold CV                       \\ \hline

\end{tabular}%
}
\begin{tablenotes}%
\item[$^{1}$] FG: Functional Genomics; FI: Functional Integration; GV: Genomic variation; Ob: Ontology-based; Nb: Network-based.
\item[$^{2}$] SVM: SVM-based; Rb: Regression-based; Tb: Tree-based; NN: Neural network; Pm: Probabilistic methods; So: Supervised learning (others); DL: Deep learning; UL: Unsupervised learning; EA: evolutionary algorithm
\item[$^{3}$] Acc: accuracy; F1: F1-measure; MCC: Matthew's correlation coefficient; Pre: precision; PRC: area under the Precision-Recall curve; ROC: area under the ROC curve; Rec: recall;  
\end{tablenotes}
\end{table*}

% %\begin{methods}
\section{Search methodology}
\label{sec:methodology}

Two main bibliographic repositories, PubMed and DBLP, were searched using keywords related to ``cancer driver genes", ``prediction", and ``machine learning". % { }as terms of interest for our survey. % While the former is most focused on biomedical literature, the latter is specialized in computer science bibliography. These repositories were chosen because, together, they cover a wide range of papers from high-impact international scientific journals and conferences in both the fields of medicine and computer science. The query for the PubMed database was constructed using keywords related to ``cancer driver genes", ``prediction", and ``machine learning"{ }as terms of interest for our survey. For DBLP, the search was conducted using only terms related to the domain of interest since all indexed articles are within the computer science field. These terms included ``cancer driver genes", ``cancer", ``disease-associated genes", and ``oncogenes". 
Our initial search returned 355 and 84 related papers in PubMed and DBLP, resulting in 420 papers after duplicates removal. Each paper was individually evaluated regarding its relevance to the survey's research topic. As inclusion criteria, we only considered papers explicitly mentioning ML as part of their methodological approach. %All titles and abstracts were screened to identify papers that employed machine learning techniques to predict cancer driver genes. When necessary, we carried out a diagonal reading of the papers to confirm their suitability for this survey's scope. 
After this initial analysis, 52 papers were considered eligible. Interestingly, many discarded papers focused exclusively on network-based methods to identify CDGs. %Here, we only selected those that addressed the use of ML techniques in their methodology in conjunction with network-based approaches. 
After in-depth analysis of papers' methodology, 36 were confirmed relevant for our survey. %We verified aspects such as which specific ML techniques were adopted in the proposed solution and the operating steps in which they were used. After this second round of paper examination, 36 papers were confirmed to be relevant for our survey, as they applied ML algorithms as a central part of the proposed computational solution for cancer drive gene prediction. 
Finally, to guarantee that no significant contribution to the field was left behind, we manually revised the selected papers' references and their citations through Google Scholar to search for relevant works that our search strategy has not retrieved. Five additional papers were selected, totaling 41 relevant papers for our survey (Table~\ref{tab:selectedPapers}). Our search was concluded on April 1st, 2021.

% %\end{methods}

\section{Overview of selected papers}
\label{sec:overview}

\begin{figure}[!t]
    \centering
    \includegraphics[width=0.48\textwidth]{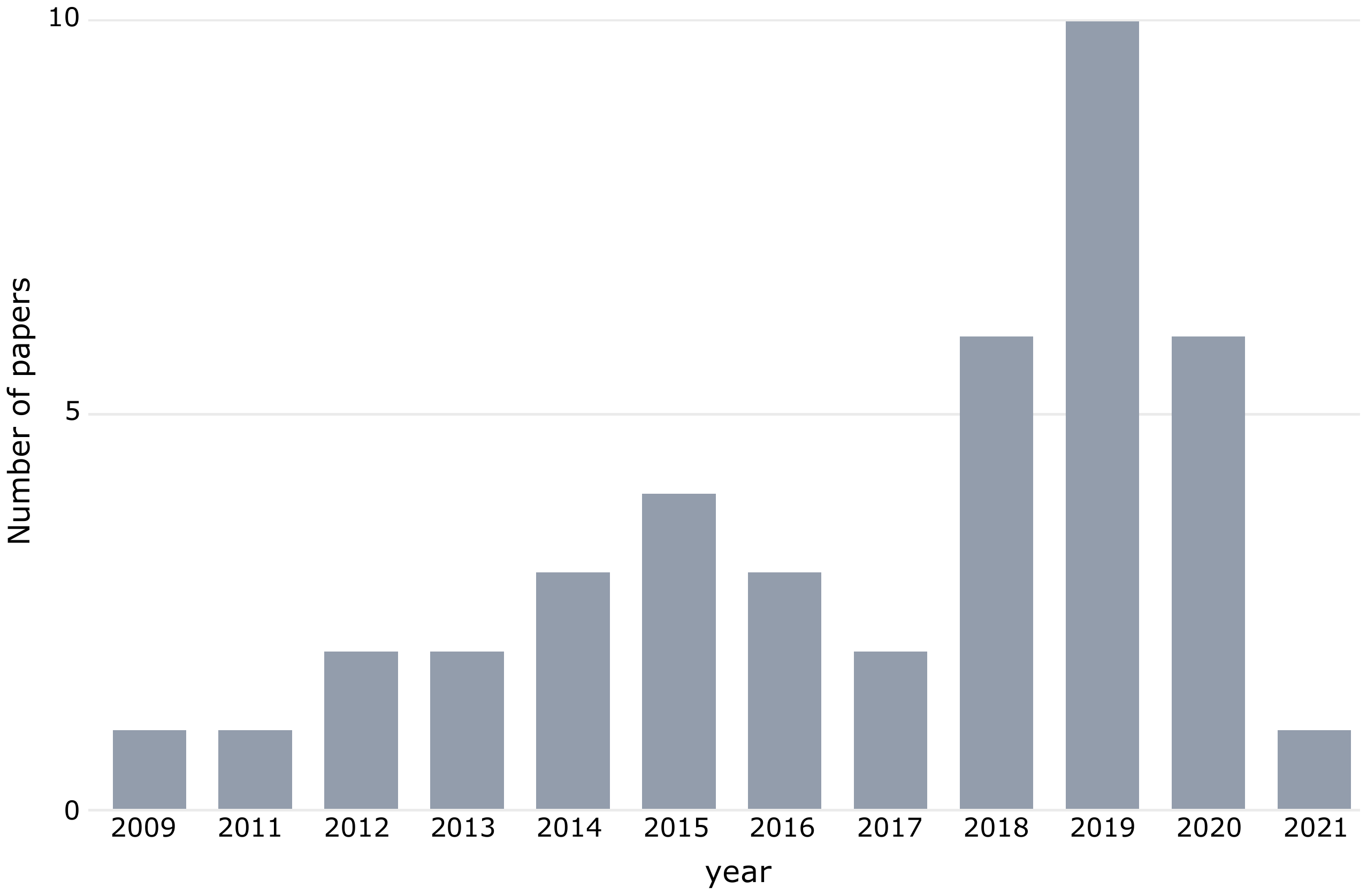}
    \caption{Number of selected papers per year of publication.}
    \label{fig:papers-year}
\end{figure}

Fig.~\ref{fig:papers-year} shows the distribution of the papers according to the publication year, suggesting a growing interest in the research topic. Most papers (56.09\%) were published after 2018 and the highest number of publications were found in 2019 (10). The low number of related papers published in 2021 is justified by the short period covered by our analysis. Papers were published mainly in scientific journals, with only five \cite{Fu2012, Tavanaei2017, Nicora2019, Schulte-Sasse2019, Cutigi2020} appearing in conference proceedings. The top three journals in terms of number of papers were Bioinformatics (6), Journal of Computational Biology (3), and Plos One (3). %Other represented journals were BMC Genomics, Genome Medicine, Nucleic Acids Research, and PLOS Computational Biology, with two publications each.

In terms of target prediction problem, 32 papers (\ie 78.04\%) concentrated in running predictions at the gene level (Supplementary Table). Among these, six papers \cite{Davoli2013, Gnad2015, Tavanaei2017, Chandrashekar2020, Colaprico2020, Lyu2020} aimed to distinguish oncogene and tumor suppressor gene (TSG) (\ie the two subclasses of CDGs), whereas the others focused on classifying a given gene as CDG or not. Seven papers targeted predictions on mutation level \cite{Mao2013, U2014, Soliman2015, Agajanian2018, Zhou2018, Agajanian2019, Wang2020}, most of which restricted the analysis for missense mutations. We also found one paper aiming at identifying cancer modules to discover cancer driver genes \cite{Manolakos2014} and other focusing on the prediction of false positive CDGs \cite{Cutigi2020}. Also, while most papers addressed cancer drivers in general, some focused on specific types of cancer, such as colon adenocarcinoma \cite{Fu2012, Luo2019}, breast cancer \cite{Celli2018, Lu2018, Luo2019}, lung adenocarcinoma \cite{Luo2019}, thyroid \cite{Celli2018}, and kidney \cite{Celli2018}. We also observed predictive models for cancer-related mutations in human protein kinases \cite{U2014} and, more specifically, in the Epidermal Growth Factor Receptor \cite{Anoosha2015}.

\begin{table*}[!hb]
\caption{Data categories and subcategories adopted in the current survey to classify selected papers according to types of features employed.\label{tab:features}}
\centering
\resizebox{\textwidth}{!}{%
\begin{tabular}{L{3cm}L{3.5cm}L{15cm}}
\hline
\multicolumn{1}{c}{Category}                           & \multicolumn{1}{c}{Subcategory} & \multicolumn{1}{c}{Description}                                                          \\ \hline
\multicolumn{1}{c}{\multirow{3}{*}{Genomic Variation}} & Mutations                       & properties of somatic single nucleotide variants (SNV) and frameshift insertion/deletion (\eg estimate of the mutation frequency, mutation ratio, mutation hotspots) \\ \cline{2-3} 
\multicolumn{1}{c}{}                                   & Copy number alteration (CNA)    & information related to amplifications, deletions, and duplication of segments of DNA that changes the number of copies of a particular DNA segment within the genome (\eg deletion and amplification frequency)                                      
\\ \cline{2-3} 
\multicolumn{1}{c}{}                                   & DNA sequence                    & raw nucleotide sequence                                                                  \\ \hline
\multirow{3}{*}{Functional Impact}                     & Functional impact scores        & outputs of \textit{in silico} variant effect predictors concerning the probability of deleterious changes on protein function                                            \\ \cline{2-3} 
                                                       & Protein-based                   & properties of protein sequence and structure, from amino acids to protein's tertiary structure      \\ \cline{2-3} 
                                                       & Evolution-based                 & evolutionary conservation scores, amino acids substitution rate, gene age, gene damage index, number of human paralogs, etc.                                             \\ \hline
\multirow{3}{*}{Functional Genomics}                   & Transcriptomics                 & large-scale gene expression profiles or statistics  derived  from  differential gene expression analyses   \\ \cline{2-3} 
                                                       & Epigenomics                     & DNA methylation, histone modifications, chromatin accessibility, DNA replication time    \\ \cline{2-3} 
                                                       & Proteomics                      & protein expression data, mainly expressed as categorical features that indicate whether or not a protein is expressed in a given human tissue                                                            \\ \hline
Ontology-based                                         & -                               & functional, cellular, or phenotypic annotations obtained from bioinformatics databases or related works                                                  \\ \hline
Network-based                                          & -                               & node properties from structural analysis of molecular networks (\eg PPI or gene-gene networks)         \\ \hline
\end{tabular}%
}
\end{table*}

To better map the current state-of-the-art in terms of available resources and adopted methodologies, two main analyses were made. The first one focused on the data types used as model's feature, while the second focused on summarizing the computational aspects of the proposed predictive model, such as type of learning and specific ML algorithms adopted. The next sections describe and organize the selected papers in terms of data categories and computational strategies.

\section{Data categories}

A core component for ML-based predictive models is the training data due to the common sense that much of the model's success depends on the fed data. A given prediction problem may have its concept represented in many different ways, which is especially true in the genomics domain. Some of these representations may be better than others in revealing the patterns we sought to learn. Thus, a clear comprehension of the possible types of features for instances representation within our domain may provide insights into current limitations and new analytical opportunities. %a better understanding of the prediction problem. %is useful to better understand the prediction problem, and identify potential limitations and opportunities.
We observed a large variety of information used as models' input features for cancer drivers prediction. After careful analysis of the selected papers, we classified them into five categories based on the properties evaluated as predictors (Table \ref{tab:features}): (1) \textit{Genomic Variation}, (2) \textit{Functional Impact}, (3) \textit{Functional Genomics}, (4) \textit{Network-based}, and (5) \textit{Ontology-based}. Subcategories were defined for some categories to better organize the corresponding features based on their semantics.

\begin{figure*}[!th]
    \centering
    \includegraphics[width=\textwidth]{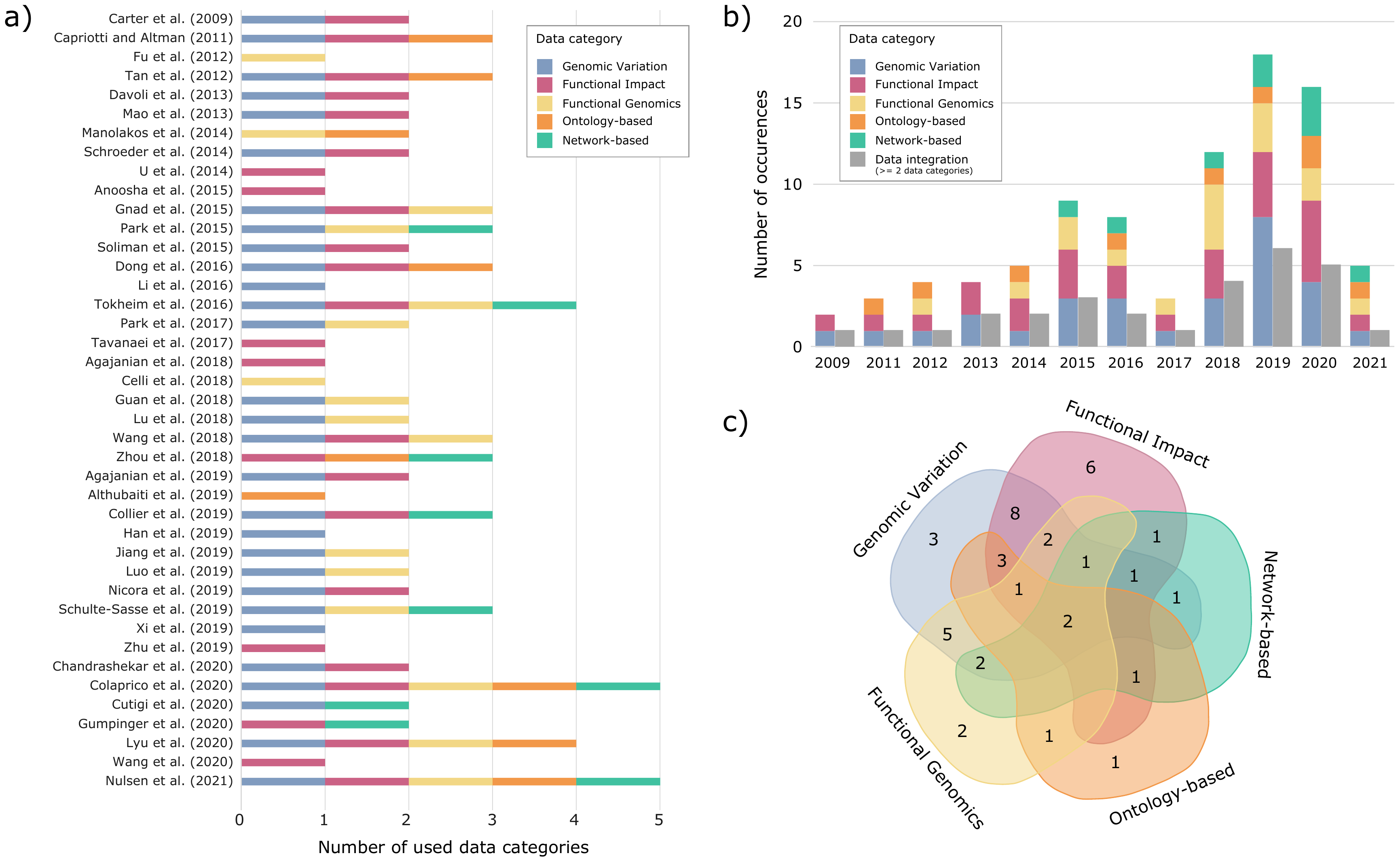}
    \caption{Analysis of data categories used as model features by selected papers. a) Relation of data categories per paper, organized by year of publication in ascending order. Each data category is represented by a distinct color, which is maintained the same across all figures. b) Distribution observed in the use of the data categories per year of publication. Papers that use more than one data category are accounted for each category, thus, the total count per year may exceed the number of papers published in the given year. Papers using two or more categories were also classified as \textit{Data integration} and are shown in the gray bar.  c) Venn diagram showing the intersection of studies in terms of employed data categories.}
    \label{fig:data-analysis}
\end{figure*}

A general overview of papers in terms of the data categories employed is shown in Fig.~\ref{fig:data-analysis}. %Fig.~\ref{fig:data-analysis}-a presents the number of data categories used by each paper, organized by year of publication. % Each data category is represented by a distinct color, which is maintained the same across the several figures. 
According to Fig.~\ref{fig:data-analysis}-a, 
\textit{Genomic variation} was the most common feature category used (29 papers, \ie 70.73\%), followed by \textit{Functional Impact} (26 papers) and \textit{Functional Genomics} (16 papers). \textit{Network-based} and \textit{Ontology-based} features were each employed in 9 papers. %For some data categories, more than one type of dataset may be used by the same paper according to the defined subcategories (Table \ref{tab:features}). 
Since some data categories contain multiple subcategories, we analyzed the number of distinct feature datasets adopted per paper (Supplementary Table) and found that it  varied from 1  to 8 (\cite{Nulsen2021}), with an average of 2.87 ($\pm$ 1.17) datasets. 

%When analyzing data categories in Fig.~\ref{fig:data-analysis}-a, 
One of our main findings regarding the features aspect of ML models is that many works adopted more than one data category. % within their computational solution.
Twenty-nine papers (\ie 70.73\%) used two or more data categories and were classified in a \textit{Data integration} category. The distribution of the number of occurrences of data categories by years of publication does not suggest any association between these factors (Fig.~\ref{fig:data-analysis}-b). 
%In Fig.~\ref{fig:data-analysis}-b, we present the distribution of the number of occurrences per data category across the years of publication, which was defined as the number of papers reporting its usage. Papers that use more than one data category are accounted for each category. %Thus, the total count per year may exceed the number of papers published in the given year (Fig.~\ref{fig:papers-year}). 
%The number of papers classified as \textit{Data integration} is shown in the right gray bar. We did not observe any association between the number of data categories used and the paper's publication year. 
Moreover, integrating different data categories is not necessarily a recent tendency, as papers published in 2009 and 2011 already proposed such strategy. Nonetheless, most of the papers using features from three or more data categories were published from 2015, and those that used four \cite{Tokheim2016, Lyu2020} or five \cite{Colaprico2020, Nulsen2021} data categories were published mainly in 2020 and 2021, which may indicate a trend for increasing data diversity in newer models. %We also observed that among the selected papers, \textit{Network-based} features were first integrated into ML-based methods in 2015 \cite{Park2015}.

The Venn diagram in Fig.~\ref{fig:data-analysis}-c summarizes the intersections among distinct data categories. The most recurrent combination was the integration of \textit{Genomic Variation} and \textit{Functional Impact} (8 papers), followed by \textit{Genomic Variation} and \textit{Functional Genomics} (5 papers). Also, while \textit{Functional Impact} was the most common data category used as a single source of features in the proposed models, the \textit{Network-based} category was not exclusively used by any ML model among the revised papers. In what follows, we %discuss in more detail the definition of each data category and 
review the particularities of each data category. % and their corresponding selected papers.

\subsection{\textbf{Genomic Variation}}

\textit{Genomic Variation} was the most frequent data category among selected papers (Fig.~\ref{fig:data-analysis}). Three papers \cite{LI2016, Han2019, Xi2019} used \textit{Genomic Variation} data as a single source of feature, whereas 23 (\ie 56.09\%) used it in combination with other categories. A long-standing hypothesis in the discovery of cancer drivers is that driver genes are mutated more frequently than expected as compared to a background mutation rate (BMR) estimated from cancer samples for a given cancer type \cite{tamborero2013comprehensive}. This hypothesis %relates to the assumption 
assumes that a true driver mutation confers a growth advantage to a tumor cell and is therefore positively selected, increasing the chance of detecting it across multiple cancer samples and occasionally among distinct cancer types \cite{chen2015deciphering}. Nonetheless, the mutational landscape of cancer consists of `mountains' of very frequently mutated genes and of `hills' of significantly but less frequently mutated genes \cite{vogelstein2013cancer}. Thus, simply characterizing candidate driver genes with a frequency-based method poses the challenge of robust estimation of the BMR: a low BMR may lead to many spurious findings, whereas a high BMR may miss the driver genes mutated at very low frequency. Using ML-based methods with \textit{Genomic Variation} data aims to allow the algorithm to learn the patterns related to driver genes and mutations %or the mutational landscape of CDGs 
from the training data. %In other words, the analysis of several properties related to gene mutation signatures with ML methods may help to identify true CDGs, including those related to lower frequency events.
%Within \textit{Genomic Variation}, 
Here, we considered three subcategories (Table~\ref{tab:features}): i) mutations, %mainly represented by somatic, nonsense, and missense single nucleotide variants (SNV) and frameshift insertion/deletions; 
ii) copy number alteration (CNA); % amplifications, deletions, and duplication of segments of DNA that changes the number of copies of a particular DNA segment within the genome; 
and iii) raw DNA sequence, which was included since information regarding DNA variants would be implicitly available. %used by the model. 

We observed that information describing mutations' properties was the most common type of feature, employed by 26 papers (\ie 63.41\%). Fifteen papers explored mutations' properties as the single source of genomic variation features, and one paper \cite{Han2019} exclusively relied on mutation-based features for model development. Features varied from a simple estimation of mutation frequency and density, used by the vast majority of papers within this subcategory, to a more complete description of a mutation's environment. %The mutation frequency is defined relative to the total number of all mutations within a gene over all tumor samples, and is often normalized by BMR or coding sequence length. In the selected papers,
Mutation frequency was mostly estimated from databases such as Catalogue of somatic mutations in cancer (COSMIC) \cite{cosmic}, The Cancer Genome Atlas (TCGA), and HapMap \cite{hapmap}. Some works \cite{Davoli2013, Gnad2015, Lyu2020} distinguished between potentially high (HiFI) or low (LoFI) functional impact mutations when computing mutation frequency, adopting predictors such as PolyPhen-2 \cite{PolyPhen-2} or MutationAssessor \cite{MutationAssessor} to estimate the mutation's functional impact. Moreover, silent mutations and LoFI missense mutations, or a combination of both \cite{Gnad2015,Lyu2020}, were often taken as a measure for a gene's BMR.

Davoli \etal \cite{Davoli2013} proposed an entropy-based mutation selection score that reflects the spatial distribution of these features, measuring the preferred occurrence of specific point mutations within a gene, termed `mutation hotspots'. This measure of positional clustering was further adopted by several papers \cite{Gnad2015, Tokheim2016, Collier2019, Luo2019, Colaprico2020, Nulsen2021}. Mutation hotspots were also detected using scores computed by OncoDriveCLUST \cite{tamborero2013oncodriveclust,Schroeder2014,Nulsen2021} and applying density estimates to aggregate closely-spaced missense mutations into peaks and compute mutation fraction inside the highest peak \cite{Chandrashekar2020}.
%Also aiming at detecting mutation hotspots, Chandrashekar \etal \cite{Chandrashekar2020} proposed to apply density estimates to aggregate closely-spaced missense mutations into peaks and then calculate mutations fraction inside these peaks and, more specifically, inside the highest peak. In addition, Nulsen \etal \cite{Nulsen2021} and Schroeder \etal \cite{Schroeder2014} adopted an estimate of hotspot mutation  %as a feature of their model 
%based on scores computed by OncoDriveCLUST \cite{tamborero2013oncodriveclust}. 
The normalized number of SNPs in the exon where the mutation is located \cite{Mao2013}, mutations' distance to closest Transcribed Sequence Start (TSS) and closest Transcribed Sequence End (TSE) \cite{Agajanian2019}, and a gene-level binary matrix summarizing mutation occurrence across all samples \cite{LI2016,Xi2019} were also adopted. 

Fourteen papers (\ie 34.15\%) \cite{Davoli2013, Schroeder2014, Gnad2015, Park2015, Dong2016, LI2016, Park2017, Guan2018, Lu2018, Jiang2019, Schulte-Sasse2019, Xi2019, Lyu2020, Nulsen2021} adopted CNA data as models' features, which was first introduced in this domain by Davoli \etal \cite{Davoli2013}. 
%The second group of genomic variation features, related to CNA data, was adopted in 14 papers (\ie 34.15\%) \cite{Davoli2013, Schroeder2014, Gnad2015, Park2015, Dong2016, LI2016, Park2017, Guan2018, Lu2018, Jiang2019, Schulte-Sasse2019, Xi2019, Lyu2020, Nulsen2021} and was first introduced as a predictive feature in this domain by Davoli \etal \cite{Davoli2013}. 
Their model analyzed the deletion and amplification frequency to distinguish among neutral genes, oncogenes, and TSGs. %Most papers within this data subcategory estimate CNA frequency by averaging CNA signal across all samples of a given cancer type. 
Two papers \cite{Gnad2015, LI2016} employed the Genomic Identification of Significant Targets In Cancer (GISTIC) \cite{gistic, gistic2} algorithm to obtain DNA amplification and deletion regions. %In Li \etal \cite{LI2016}, authors created a binary matrix defining the status of each gene in all samples,  where each cell has the value of 1 if one of the following two conditions is satisfied: the mutation status of the gene is classified as somatic or the gene is located in statistically significant variable regions according to GISTIC estimates. 
Gnad \etal \cite{Gnad2015} used as features the sums of amplification or deletion frequencies across 13 TCGA cancer types. We note that 11 out of the 14 papers that fall within the CNA subcategory also adopted mutation data. However, in most cases, the features from the different subcategories were used independently rather than combined into a single input feature. One exception is the work by Schulte-Sasse \etal \cite{Schulte-Sasse2019}, in which the sum of SNVs and CNA is averaged across all samples of a given cancer type to estimate a cancer mutation rate per gene. Furthermore, we observed that from the three papers that apply only CNA features from the \textit{Genomic variation} category, two \cite{Park2017, Guan2018} also adopted gene expression data in their analyses, since CNA is known to mediate phenotypic changes through their impact on expression. Park \etal \cite{Park2017} quantified the association between CNA and gene expression using a previously proposed method \cite{yuan2011sparse}, differentiating between cis-effects (\ie when gene expression is influenced by CNA in proximal genes within a several Mb window) and trans-effects (\ie  when gene expression is influenced by remote alterations throughout the genome). Guan \etal \cite{Guan2018} used gene-level summarized cell-lines CNA and expression profiles obtained from the Cancer Cell Line Encyclopedia (CCLE). 

Finally, only one paper \cite{Agajanian2019} used raw DNA sequence as model's features. 
Mutations were represented by stacking two nucleotide sequences on top of each other, one with the original nucleotide for a given position and the other with the mutated version, and a convolutional neural network (CNN) was trained to automatically extract driver-related sequence patterns.

\subsection{\textbf{Functional Impact}}

The functional impact (FI) of SNVs on protein function was analyzed by 63.41\% of papers. Their goal is to better identify lowly recurrent mutated driver genes or driver genes that are mutated late during tumor development, which are more challenging cases for methods that exclusively analyze genomic variation features \cite{gonzalez2012functional}. This has become a widely applied strategy even outside the scope of ML-based papers, as reflected by the several computational methods developed for predicting functionally damaging effects of SNVs %occuring either in protein coding regions and non-coding regions 
\cite{chen2020comprehensive, wu2016dbwgfp}. 
Three subcategories were created (Table~\ref{tab:features}): i) functional impact scores provided by \textit{in silico} predictors for variant effect; %and reflect the probability that a given SNV will cause a functional change in the protein function; 
ii) protein-based features; %given by properties related to protein sequence and structure; 
and iii) evolution-based features. %, measuring evolutionary conservation of nucleotides or amino acids positions. 
Although most FI predictors rely on features related to proteins' properties or to evolutionary conservation \cite{chen2020comprehensive}, we did not consider this in our classification. %. Nonetheless, we did not consider their FI predictors' features within our classification, 
Papers that only employed \textit{in silico} FI analysis were placed under the \textit{functional impact scores} subcategory despite implicitly using information from the other two subcategories.

% \usepackage{graphicx}
\begin{table*}[!ht]
\caption{\textit{In silico} tools for functional impact prediction adopted by papers included in our survey \label{tab:fi_tools}}
\centering
\resizebox{\textwidth}{!}{%
\begin{tabular}{L{4.7cm}L{17cm}}
\hline
\textbf{Reference} & \textbf{Tools} \\ \hline
Capriotti and Altman (2011) \cite{Capriotti2011} & PANTHER \cite{panther} \\ \hline
Davoli \etal (2013) \cite{Davoli2013}  & Polyphen-2 \cite{PolyPhen-2}\\ \hline
Mao \etal (2013) \cite{Mao2013} & SIFT \cite{ng2003sift}, PolyPhen-2 \cite{PolyPhen-2}, CONDEL \cite{condel}, MutationAssessor \cite{MutationAssessor}, PhyloP \cite{PhyloP}, GERP++ \cite{GERP++}, and LRT \cite{LRT} \\ \hline
Schroeder \etal (2014) \cite{Schroeder2014}  & OncodriveFM \cite{Oncodrive-fm}\\ \hline
Gnad \etal (2015) \cite{Gnad2015} & MutationAssessor \cite{MutationAssessor} \\ \hline
Dong \etal (2016) \cite{Dong2016}  & SIFT \cite{ng2003sift}, PolyPhen-2 \cite{PolyPhen-2}, LRT \cite{LRT}, MutationTaster \cite{schwarz2010mutationtaster}, MutationAssessor \cite{MutationAssessor}, FATHMM \cite{shihab2013predicting}, GERP++ \cite{GERP++}, PhyloP \cite{PhyloP}, CADD \cite{kircher2014general}, VEST \cite{carter2013identifying}, SiPhy \cite{garber2009identifying}, FunSeq2 \cite{funseq2}\\ \hline
Tokheim \etal (2016) \cite{Tokheim2016} & VEST \cite{carter2013identifying} \\ \hline
Agajanian \etal (2018) \cite{Agajanian2018}  & SIFT \cite{ng2003sift}, PolyPhen-2 \cite{PolyPhen-2}, LRT \cite{LRT}, MutationAssessor \cite{MutationAssessor}, MutationTaster \cite{schwarz2010mutationtaster}, FATHMM \cite{shihab2013predicting}, MSRV \cite{jiang2007sequence}, SinBaD \cite{lehmann2013exploring}, GERP++ \cite{GERP++}, SiPhy \cite{garber2009identifying}, PhyloP \cite{PhyloP}, Grantham \cite{grantham1974amino}, CADD \cite{kircher2014general}, GWAVA \cite{ritchie2014functional}, MetaLR \cite{dong2015comparison}, and MetaSVM \cite{dong2015comparison}\\ \hline
Wang \etal (2018) \cite{Wang2018}  & SIFT \cite{ng2003sift} and GERP++ \cite{GERP++} \\ \hline
Zhou \etal (2018) \cite{Zhou2018}   & ENTPRISE \cite{zhou2016entprise}  \\ \hline
Agajanian \etal (2019) \cite{Agajanian2019} & SIFT \cite{ng2003sift}, PolyPhen-2 \cite{PolyPhen-2}, LRT \cite{LRT}, MutationAssessor \cite{MutationAssessor}, MutationTaster \cite{schwarz2010mutationtaster}, FATHMM \cite{shihab2013predicting}, MSRV \cite{jiang2007sequence}, SinBaD \cite{lehmann2013exploring}, GERP++ \cite{GERP++}, SiPhy \cite{garber2009identifying}, PhyloP \cite{PhyloP}, Grantham \cite{grantham1974amino}, CADD \cite{kircher2014general}, GWAVA \cite{ritchie2014functional}, MetaLR \cite{dong2015comparison}, and MetaSVM \cite{dong2015comparison}\\ \hline
Collier \etal (2019) \cite{Collier2019} & Polyphen-2 \cite{PolyPhen-2}\\ \hline
Zhu \etal (2019) \cite{Zhu2019} & 20/20+ \cite{Tokheim2016}, MutSigCV \cite{lawrence2013mutational}, OncodriveFM \cite{Oncodrive-fm}, OncodriveCLUST  \cite{tamborero2013oncodriveclust}, DrGaP \cite{hua2013drgap}, and MUFFINN \cite{cho2016muffinn}\\ \hline
Colaprico \etal (2020) \cite{Colaprico2020}  & VEST \cite{carter2013identifying} \\ \hline
Gumpinger \etal (2020) \cite{Gumpinger2020} & MutsigCV  \cite{lawrence2013mutational} \\ \hline
Lyu \etal (2020) \cite{Lyu2020}  & VEST \cite{carter2013identifying}, PolyPhen-2 \cite{PolyPhen-2}\\ \hline
Wang \etal (2020) \cite{Wang2020} & GERP++ \cite{GERP++}, PhastCons, PhyloP \cite{PhyloP}, LRT \cite{LRT}, SiPhy \cite{garber2009identifying}, FATHMM \cite{shihab2013predicting}, fitCons \cite{gulko2015method}, MutationAssessor \cite{MutationAssessor}, MutationTaster \cite{schwarz2010mutationtaster}, PolyPhen2-HDIV \cite{PolyPhen-2}, PolyPhen2-HVAR \cite{PolyPhen-2}, PROVEAN \cite{choi2015provean}, SIFT \cite{ng2003sift}, VEST3 \cite{carter2013identifying}, CADD \cite{kircher2014general}, DANN \cite{quang2015dann}, Eigen \cite{ionita2016spectral}, FATHMM-MKL \cite{shihab2015integrative}, GenoCanyon \cite{lu2015statistical}, M-CAP \cite{jagadeesh2016m}, MetaLR \cite{dong2015comparison}, MetaSVM \cite{dong2015comparison}, and REVEL \cite{ioannidis2016revel} \\ \hline
\end{tabular}%
}
\end{table*}

The largest subcategory was FI scores, used by 18 papers. A summary of adopted FI predictors is given in Table~\ref{tab:fi_tools}. Most works used the tool's predicted scores and/or p-values as models' features, while others (\eg \cite{Collier2019}) estimated the number of damaging missense mutations per gene using a pre-defined score threshold. Several works integrated the output of multiple FI predictors in their feature vectors \cite{Mao2013, Dong2016, Agajanian2018, Agajanian2019, Zhu2019, Wang2020}. Dong \etal \cite{Dong2016} used 11 tools for point coding mutations and one tool, FunSeq2 \cite{funseq2}, to annotate noncoding variants. Agajanian and colleagues \cite{Agajanian2018, Agajanian2019} built a feature vector with 32 prediction scores and categories provided by 16 algorithms. Zhu \etal \cite{Zhu2019} identified the consensus driver genes using six tools with complementary strategies, including a ML-based tool (\ie 20/20+ \cite{Tokheim2016}).  Wang \etal \cite{Wang2020} explored the largest diversity of tools, adopting 23 algorithms that included conservation-based, ensemble-based, and function-prediction methods.

%Regarding the protein-based subcategory, features varied from the analysis of individual amino acid properties to physicochemical properties of the complete protein tertiary structure, and were  adopted by 11 papers.
In the protein-based subcategory, adopted by 11 papers, some works \cite{Carter2009, Tan2012, Mao2013} evaluated the changes in residues' charge, volume, polarity, the hydrophobicity resulting from the mutation, the predicted residue solvent accessibility and backbone flexibility, the mutation's effect on protein stability, and the probability that the secondary structure of the wild type residue's region is helix, loop, or strand. Anoosha \etal \cite{Anoosha2015} considered 49 physical, chemical, energetic, and conformational parameters comparing wild type and mutant residues, as well as neighboring residue information at different window lengths. Feature vectors encoding the mutated sequence and the mutation's local sequence environment \cite{Capriotti2011} or the biochemical properties of the atomic coordinates of proteins' PDB structure \cite{Tavanaei2017} were also proposed. Other characteristics considered were the fraction of the affected protein structure \cite{Zhou2018}, amino acid composition of the residues in contact with the mutation or of its domain \cite{Zhou2018}, amino acid's position in the codon or protein \cite{Agajanian2019}, number of complexes the protein participates \cite{Nulsen2021}, and background probability for observing the wild type or mutant residue in the first, middle, or last position of an amino acid triple, or at the center of a window of 5 amino-acid residues \cite{Carter2009, Tan2012}. Finally, amino acids substitution scores were employed by several studies \cite{Carter2009, Tan2012, Mao2013, U2014, Anoosha2015}, most of which integrated distinct substitution scoring matrices. Tan \etal \cite{Tan2012}, for instance, defined 51 features by integrating dozens of substitution scoring matrices from the AAIndex database, which was explored in other studies \cite{Anoosha2015,Soliman2015}.

The evolution-based subcategory was employed in 14 studies, most of which computed evolutionary conservation scores using distinct strategies or tools \cite{Carter2009, Capriotti2011, Mao2013, U2014, Anoosha2015, Tokheim2016, Agajanian2018, Zhou2018, Agajanian2019, Colaprico2020, Lyu2020}. MGAentropy, for instance, was employed by three papers \cite{Mao2013,Tokheim2016, Lyu2020}. The rationale is that the more an amino-acid residue is functionally or structurally important, the more it is conserved over evolution. %The MGAentropy, which measures the multispecies conservation of missense mutation sites, was employed by three papers \cite{Mao2013,Tokheim2016, Lyu2020}. 
Chandrashekar \etal \cite{Chandrashekar2020} computed the mean substitution rate of all protein's positions as well as of positions under the highest peak of closely-spaced mutations. Lyu \etal \cite{Lyu2020} employed the gene age, % (\ie time of evolutionary origin based on the presence/absence of orthologs in vertebrates), 
gene damage index, and the number of human paralogs for each gene, among others. Finally, Nulsen \etal \cite{Nulsen2021} included the information of genes' evolutionary origin (\ie pre-metazoan, metazoan, vertebrate, or post-vertebrate). % using a set of four binary features.

%Based on an overall analysis of papers classified in the \textit{Functional Impact} category, 
We observed that six studies \cite{U2014, Anoosha2015, Tavanaei2017, Agajanian2018, Zhu2019, Wang2020} classified in the \textit{Functional Impact} category only used this data type for developing their predictive models (Fig.~\ref{fig:data-analysis}-a). Among these, Zhu \etal \cite{Zhu2019} and Wang \etal \cite{Wang2020} defined their features exclusively based on scores from FI predictors. This is not surprising, as this type of feature carries a very rich underlying information provided by the manifold properties analyzed by each FI predictor. 
\subsection{\textbf{Functional Genomics}}

About 39\% of papers used features related to \textit{Functional Genomics}, which we divided in three subcategories (Table~\ref{tab:features}): i) transcriptomics; ii) epigenomics; and iii) proteomics. The main motivation %for using \textit{Functional Genomics} data 
comes from previous observation that mutation frequency across the genome is strongly correlated with transcriptional activity and DNA replication timing \cite{Jiang2019, lawrence2013mutational}, and that driver gene mutations are tightly tied to DNA methylation landscape in multiple types of cancer \cite{chen2017significant}. Moreover, mutation rates vary among individual genes and are influenced by many factors, including the aforementioned ones (\ie gene expression, replication timing, DNA methylation) and others such as chromatin state \cite{schuster2012chromatin}. Thus,  %integrating these types of predictors in the ML models may help improve the detection of cancer drivers.
integrating these types of predictors in the ML models may be helpful in differentiating cancer genes from the rest of human genes \cite{Nulsen2021}.

Transcriptomic data %, referring mainly to gene expression profiles or statistics derived from differential gene expression analyses, 
was employed as features in 15 papers, representing 93.75\% of papers classified as \textit{Functional Genomics}. %  In relation to the papers that were classified as \textit{Functional Genomics}, this represents 93.75\%. 
The first model to use this type of data was the one proposed by Fu \etal \cite{Fu2012}, which analyzed microarray gene expression data from 174 patients with colon adenocarcinoma cancer obtained from TCGA. This is the only work to rely exclusively on transcriptomic data for extracting patterns related to oncogenic genes. Gene expression levels were also provided as model inputs in several other works \cite{Manolakos2014, Park2015, Guan2018}. Other papers summarized gene expression profiles by computing differential gene expression scores \cite{Gnad2015, Tokheim2016, Lu2018, Wang2018, Schulte-Sasse2019, Colaprico2020, Lyu2020} or average expression in cancer tissues or cell lines \cite{Jiang2019}. In Nulsen \etal \cite{Nulsen2021}, authors adopted features quantifying the number of tissues expressing each gene, as well as binary features indicating whether the gene is expressed is a certain number of tissues (\ie in 0 tissues, in 1 to 6 tissues, in 7 to 36 tissues, or in more than 36 tissues).

Among the selected papers, we observed five (\ie. 12.2\%) \cite{Tokheim2016, Celli2018, Jiang2019, Schulte-Sasse2019, Lyu2020} using features from the epigenomic domain. Despite the low number of papers, a wide range of information was registered. In Tokheim \etal \cite{Tokheim2016}, authors used as features the DNA replication time and the 3D chromatin interaction capture (HiC) statistic, which is a measure of open vs. closed chromatin state. Both information was obtained from the MutSigCV webpage \cite{lawrence2013mutational} as provided by the Broad Institute. Jian and colleagues \cite{Jiang2019} also adopted these two features, complemented by chromatin accessibility by ATAC-Seq data and beta values obtained from DNA methylation data. For both data types, authors computed an average per gene across all subjects of the same cancer type. 

In Schulte-Sasse \etal \cite{Schulte-Sasse2019}, authors analyzed data from 16 types of cancer obtained from TCGA and represented each gene by a 16 $\times$ 3-dimensional vector, containing information about the differential DNA methylation from the gene's promoter region, the differential gene expression, and the gene's mutation rate in each type of cancer analyzed. Moreover, Lyu \etal \cite{Lyu2020} explored several types of epigenetic properties, including as features of their model early replication timing quantified by the S50 score, super enhancer annotation, promoter and gene-body cancer–normal methylation difference, and histone modifications from the ENCODE project. Finally, we note that one paper \cite{Celli2018} focused exclusively on DNA methylation data. Celli \etal \cite{Celli2018} extracted datasets of large-scale DNA methylation profiling from TCGA for three types of cancer (breast, kidney, and thyroid carcinomas) and used the beta value as predictors for their classifiers.

Finally, only one paper \cite{Nulsen2021} was classified in the proteomics subcategory for using features that reflect the number of healthy human tissues expressing a protein, and whether a protein is expressed in 0 to 8 tissues, or in 41 or more tissues. Information was obtained from the Protein Atlas v18. 
%In the third subcategory, related to proteomic data, we classified only one paper that analyzed protein expression as input features. Nulsen \etal \cite{Nulsen2021} included three features related to this data type, which were taken from the Protein Atlas v18. The first feature reflects the number of healthy human tissues expressing a protein, and the two others provide binary indicators of whether a protein is expressed in 0 to 8 tissues, or in 41 or more tissues. %Categorical protein expression data, which indicates whether or not a protein is expressed in a given human tissue, was taken from the Protein Atlas v18.
An analysis of papers using \textit{Functional Genomics} data shows that %the use of transcriptomic data was the most common type of feature from this category. Eleven of the 16 papers that fall within this category 
11 studies used only transcriptomic data, while three \cite{Jiang2019, Schulte-Sasse2019, Lyu2020} combined transcriptomics and epigenomics, and one \cite{Nulsen2021} combined transcriptomic- and proteomic-based features. Thus, except for one paper that used solely DNA methylation as input features of their prediction model \cite{Celli2018}, all other works that adopted features from the epigenomics domain also considered transcriptomic data. %According to Nulsen \etal \cite{Nulsen2021}, these system-level properties may be very useful in differentiating cancer genes from the rest of human genes.

\subsection{\textbf{Ontology-based}}

In the \textit{Ontology-based} category, %, we considered all papers that employed as input features for their models functional, cellular, or phenotypic annotations obtained from bioinformatics databases or related works. These 
annotations varied from characterizing biological processes such as those provided by Gene Ontology (GO) to more specific categorization of genes' role in organisms' functioning and phenotype (\eg essentiality, diseases involvement, cellular localization, etc.). The rationale is that prior knowledge regarding genes association with diseases or with molecular processes implicated in cancer may help improve gene prioritization in the search for CDGs. Nine papers (21.95\%) were classified within this subcategory. While most papers used ontology-based features in combination to other data types, one paper focused its analysis exclusively on public background knowledge about gene functions, cellular locations, and cellular and organism phenotypes obtained from ontology databases \cite{Althubaiti2019}. To identify CDGs and, subsequently, driver mutations, Althubaiti \etal \cite{Althubaiti2019} adopted biomedical ontologies provided by Cellular Microscopy Phenotype Ontology (CMPO), Gene Ontology (GO), and Mammalian Phenotype Ontology (MP) to learn an embedding for each gene using a neuro-symbolic approach.%. An approach based on neuro-symbolic feature learning was used to obtain an embedding for each gene encoding the known association of genes and their functions and phenotypes, which is employed as a gene's feature vector in the proposed ML model.

Information from biological processes were integrated into two frameworks \cite{Capriotti2011, Colaprico2020}, either as features encoding the number of GO terms associated to a given gene \cite{Capriotti2011} or as prior knowledge about biological process linked to cancer to identify their mediators \cite{Colaprico2020}. Prior knowledge regarding disease involvement of protein under altered function was also explored by selected works, using for instance, the databases Phenolyzer \cite{Dong2016} and GeneCards \cite{Zhou2018}. Moreover, we observed two papers adopting protein essentiality annotations as input features based on the rationale that functional changes in essential proteins are more likely to be associated with diseases \cite{Zhou2018,Nulsen2021}.

Tan \etal \cite{Tan2012} used several annotated features collected from UniProt KB, SwissProt variant page, and COSMIC database, including known motifs, known zinc fingers, mutagenic sites, and whether it refers to metal-binding, DNA binding, or transmembrane regions. 
%, as well as calcium, phosphate, disulfide, and carbohydrate annotations. In the work by Manolakos \etal \cite{Manolakos2014}, annotations related to regulatory genes were integrated into the model. The data was created based on transcription factors obtained from the HPRD database \cite{vaquerizas2009census}, and their idea is to identify these elements that play a crucial role at the molecular pathway level by influencing the expression levels of several other genes so as to detect cancer-related modules formed by regulatory genes and their downstream targets. 
Annotations about regulatory roles or interactions were integrated into the model proposed by two works \cite{Manolakos2014, Nulsen2021}. In the first \cite{Manolakos2014}, authors explored transcription factors obtained from the HPRD database \cite{vaquerizas2009census} to identify elements that play a crucial role at the molecular pathway level by influencing the expression of several other genes so as to detect cancer-related modules formed by regulatory genes and their downstream targets. In the second \cite{Nulsen2021}, authors included the number of microRNAs targeting a given gene \cite{Nulsen2021} registered at miRTarBase \cite{chou2018mirtarbase} and miRecords \cite{xiao2009mirecords} databases, motivated by the hypothesis that genes related to canonical driver proteins are targeted by more microRNAs.  Finally, one work \cite{Lyu2020} proposed as predictive features annotations regarding super enhancer from the dbSUPER database \cite{khan2016dbsuper} and cell proliferation scores as a proxy for essentiality. %and gene essentiality using cell proliferation scores as a proxy.

\subsection{\textbf{Network-based}}

The \textit{Network-based} category relates to features extracted from molecular networks and was used by nine papers. In biological systems, %it is important to observe 
the interactions between proteins are important for the comprehension of cell physiology since the vast majority of proteins interact with others for proper biological activity. Moreover, the high interconnectivity among cellular elements implies that an abnormality in a given gene or protein may spread along the links of the molecular network and impact on the activity of other elements. Thus, the hypothesis that a disease phenotype is rarely a consequence of a defect on a single gene or protein but rather of alterations in the biological processes that interact in a complex network has motivated network-based approaches to study human diseases \cite{barabasi2011network}. %In fact, when searching for related papers to this survey, we found a large number of works proposing network-based approaches to identify CDGs. According to our inclusion and exclusion criteria, we only selected those that combined ML algorithms with network-based properties. %These papers were classified as \textit{Network-based} and account for nine out of the 41 (\ie 21.95\%) selected papers.

%Protein-protein interaction (PPI) networks were a recurrent source of features in this domain. 
The predictive features in this domain were mainly related to centrality measures obtained from Protein-protein interaction (PPI) networks, such as degree, betweenness, and clustering coefficient \cite{Tokheim2016, Zhou2018, Colaprico2020, Cutigi2020, Nulsen2021}, based on the hypothesis that proteins encoded by canonical drivers tend to have higher centrality than other proteins \cite{Nulsen2021}. Cutigi \etal \cite{Cutigi2020} considered several other node properties: closeness, eigenvector, coreness, average of neighbors’ degree, leverage, information, and bridging. %Their framework integrates ten features from the genomic variation domain related to mutations, and ten node properties to characterize the importance of a node in a given network. 
Moreover, four distinct PPI networks were combined by the authors to extract node measures: ReactomeFI \cite{ReactomeFI}, HINT \cite{hint}, HPRD \cite{hprd}, and HuRI \cite{huri}. %Nulsen and colleagues \cite{Nulsen2021} also explored the clustering coefficient measure. In addition, the authors 
Nulsen \etal \cite{Nulsen2021} defined a binary feature indicating whether a protein is a hub or not in the PPI network if it occurs in the top 25\% of degree distribution, using as basis the union of BioGRID \cite{biogrid}, MIntAct \cite{mintact}, DIP \cite{dip}, and HPRD networks \cite{hprd}. 

Park \etal \cite{Park2015} %developed a new statistical %method to better understand the heterogeneous cancer system by identifying driver genes and their interactions. Their method aims 
%to perform network selection as a group selection based on an input network, and then run gene selection for each selected network to uncover potential driver genes. Their approach 
leverage gene networks constructed based on comprehensive genome-scale information, including PPIs, gene expression level, SNVs, and CNA. Collier \etal \cite{Collier2019} %proposed an approach to predict new CDGs given a list of known ones that employs an integrated kernel function to quantify the similarity of any pair of genes. The 
quantify the similarity of gene pairs using an integrated kernel function that combines prior information about mutations and PPI network. %, calculating similarity between two genes based on somatic mutation patterns or from the relative positions of genes in a PPI network. %Their results underline the fact that mutations and PPI data provide complementary information that is useful in this prediction task.
Schulte-Sasse \etal \cite{Schulte-Sasse2019} also proposed a multiomics approach for prediction of CDGs, combining PPI from the ConsensusPathDB with gene mutation rates, DNA methylation, and gene expression levels. %Their model does not explores hand-crafted features based on network structural analysis. 
Instead of exploring hand-crafted network-based features, they adopted a ML algorithm known as Graph Convolutional Network (GCN) that is able to directly analyze graph-structured data and recognize patterns in a local neighborhood of a node, using the PPI network as the input graph. %Thus, the PPI network is used as the input graph for the GCN and the other omics datasets are provided as node features. 
%A similar idea of fully exploring the PPI network without manually computing node-based centrality measures was adopted in the work by Gumpinger \etal \cite{Gumpinger2020}. Authors used the InBio Map PPI network as the representation of gene interactions, and proposed a node embedding procedure that integrates the PPI network structure with nodes' MutSig p-values.
A similar approach was proposed by Gumpinger \etal \cite{Gumpinger2020}, using the InBio Map PPI network as the representation of gene interactions to generate node embeddings that integrate the network structure with nodes' MutSig p-values. Their findings suggest that a node's context within a network introduces valuable information in the prediction of cancer drivers. 

 \begin{figure*}[!t]
    \centering
    \includegraphics[width=0.85\textwidth]{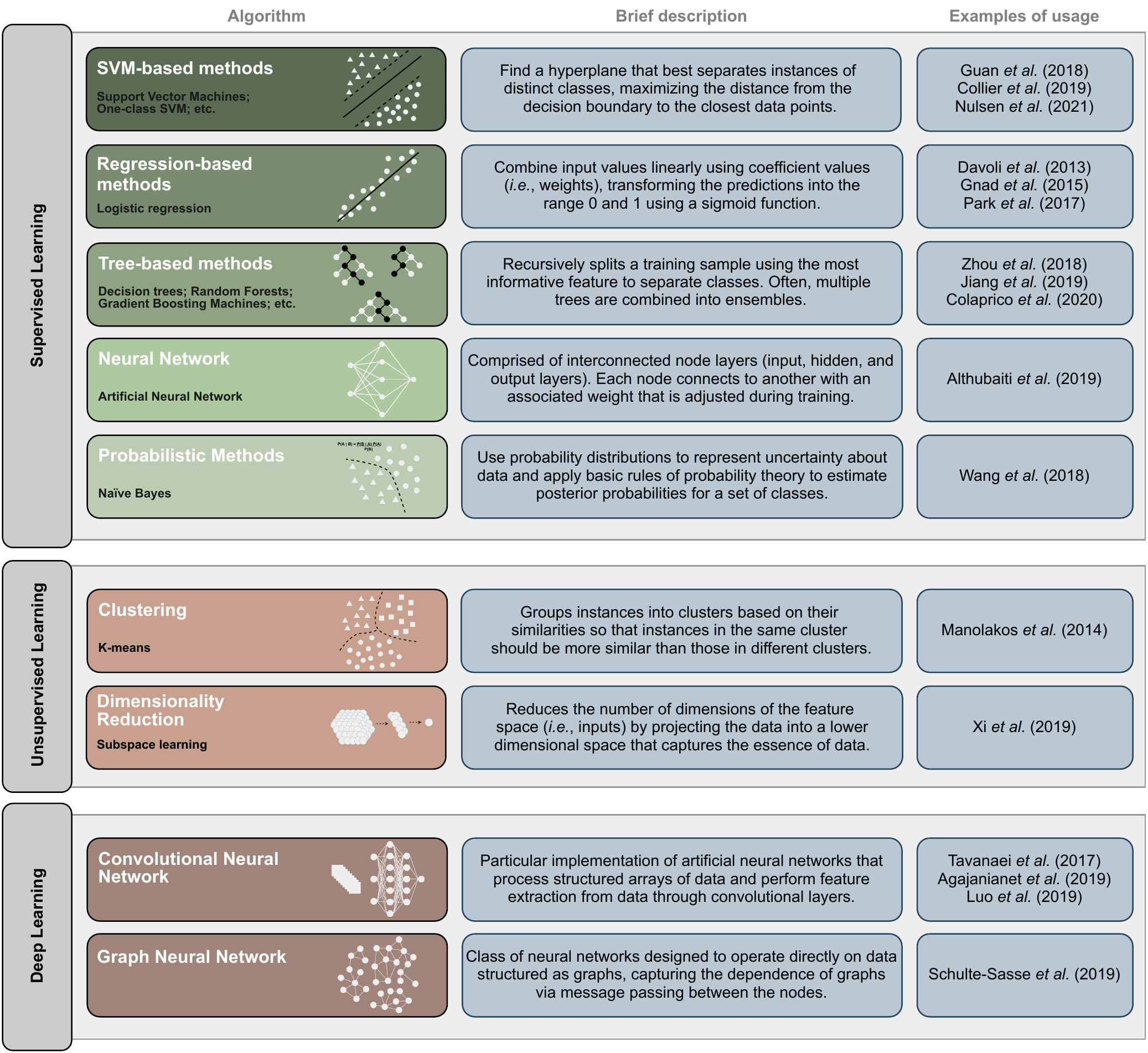}
    \caption{Classes and examples of machine learning algorithms identified through this survey with applications in cancer driver gene prediction.}
    \label{fig:algorithm-summary}
\end{figure*}

\section{Machine Learning Strategies}

 \begin{figure*}[!ht]
    \centering
    \includegraphics[width=\textwidth]{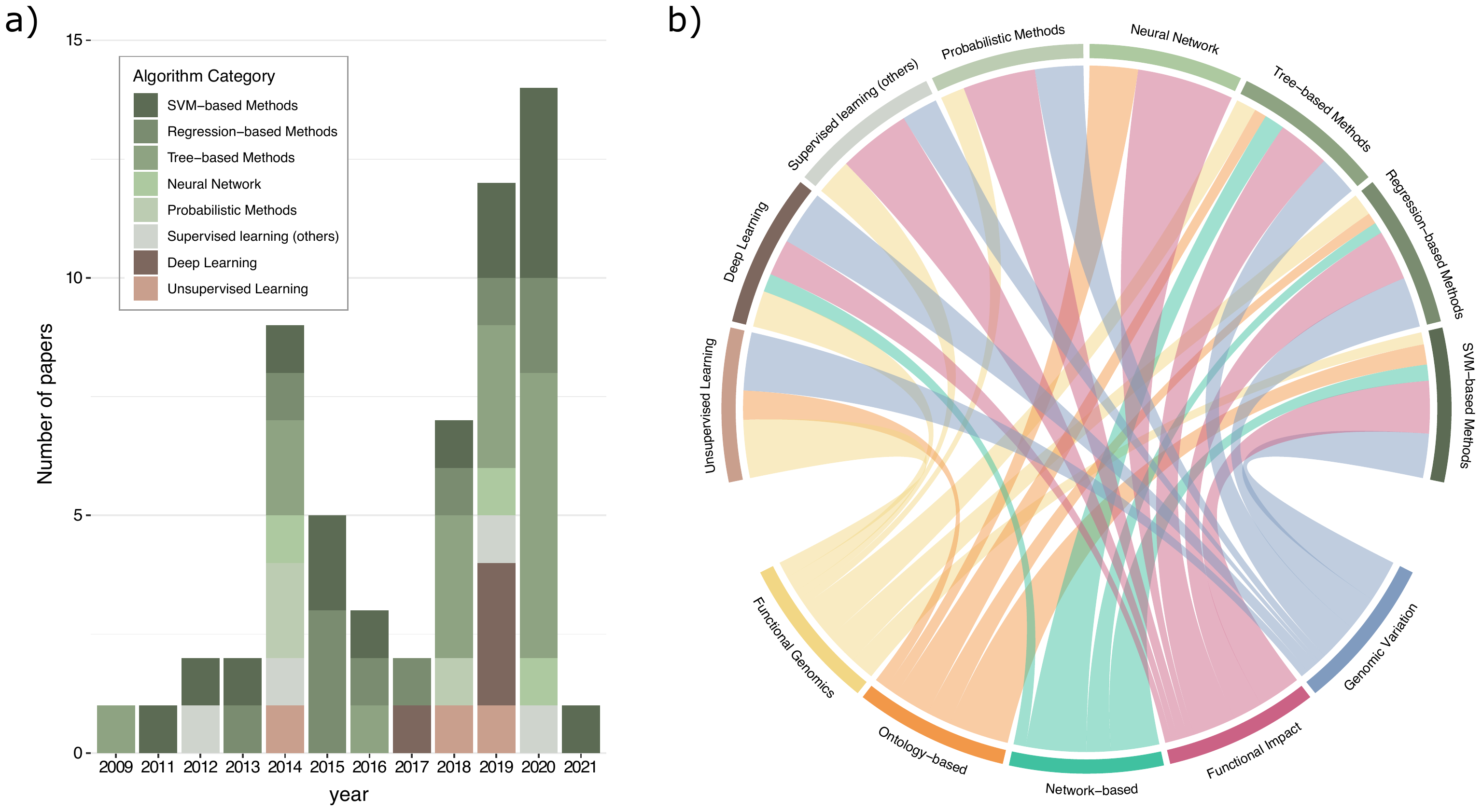}
    \caption{Analysis of types of algorithms used for model development in the selected papers. a) Distribution of the number of occurrences found for each algorithm category per year of publication. Papers that adopted more than one type of algorithm are accounted for each one of them. b) Association between data categories and algorithm categories.}
    \label{fig:algorithm-analysis}
\end{figure*}

In terms of computational methods used, we categorized the selected papers into supervised learning, unsupervised learning, and deep learning. In supervised learning, a set of labeled  instances (\ie examples) is provided in the training phase of the algorithm, which will learn to extract the underlying patterns in order to classify new data or predict the outcome of unseen instances. In unsupervised learning, no labels are provided and thus the algorithm aims to discover hidden patterns in data based only on internal knowledge, by searching for common characteristics of instances or any associations among characteristics. Finally, in deep learning, artificial neural networks with a large number of hidden layers are combined with representation learning to transform the data into different levels of abstraction and be able to learn complex patterns. An overall scheme of the algorithms used is presented in Fig.~\ref{fig:algorithm-summary}. For a gentle introduction to ML, we refer reader to the recent review by Greener \etal \cite{greener2021guide}. 
% the learning algorithms are composed by several levels of representation, in which every level use the information from the previous one to learn patterns more deeply.
We note that two papers do not fall into any of these categories: the first uses a genetic algorithm \cite{LI2016}, which is a population-based learning algorithm inspired by nature, and the second is a web-based consensus CDG caller that performs rank-based aggregation of FI \textit{in silico} predictors' output, including ML-based tools \cite{Zhu2019}. 

The vast majority of papers (\ie 80\%) modeled the problem as a supervised ML task, which is expected since known examples of true cancer drivers are available in specialized databases and may be used as training data. We also found four papers based on deep learning \cite{Tavanaei2017, Agajanian2019, Luo2019, Schulte-Sasse2019} and three papers \cite{Manolakos2014, Lu2018, Xi2019} adopting unsupervised learning techniques. 
To provide a more granular discussion about the supervised learning approaches, we further divided them into six subcategories: Support Vector Machine (SVM)-based methods, regression-based methods, tree-based methods, neural network, probabilistic methods, and other approaches that do not fall into any of the previous classes. Fig.~\ref{fig:algorithm-analysis}-a shows the algorithm categories distribution across the year of paper publication. Since some papers adopted more than one algorithm, the number of papers per year may exceed the total records (Fig.~\ref{fig:papers-year}). In Fig.~\ref{fig:algorithm-analysis}-b, we summarize the observed associations between ML strategies and data categories.

\subsection{\textbf{Methods based on supervised learning}}

Tree-based and SVM-based methods were the most common approaches among supervised learning techniques, observed in 39.02\% and 36.59\% of selected papers. The first proposals by Carter \etal \cite{Carter2009} and Capriotti \etal \cite{Capriotti2011} were based on these algorithms. Among the SVM-based approaches, whereas most papers adopted the traditional SVM algorithm \cite{Capriotti2011, Tan2012, Mao2013, U2014, Soliman2015, Dong2016, Guan2018, Cutigi2020, Gumpinger2020, Lyu2020, Wang2020}, we observed three papers using OneClass SVM \cite{Collier2019, Nicora2019, Nulsen2021} and one paper using Sequential Minimal Optimization (SMO) \cite{Anoosha2015}. SVM is a popular and consolidated technique in the field, as it continues to be largely applied throughout the years since 2011.

Considering tree-based methods, although some papers explored the traditional decision tree algorithm \cite{Schroeder2014, U2014, Wang2020}, the vast majority used tree-based ensemble classifiers. Ensemble methods train multiple weak classifiers, such as decision trees, and combine their output (\eg with majority voting) to achieve a better predictive performance. We observed a frequent use of Random Forests (RF) \cite{Carter2009, Schroeder2014, U2014, Tokheim2016, Agajanian2018, Celli2018, Agajanian2019, Nicora2019, Chandrashekar2020, Colaprico2020, Cutigi2020, Gumpinger2020, Lyu2020, Wang2020}, as well as Gradient Boosting Trees (GBT) \cite{Zhou2018, Agajanian2018}, and eXtreme Gradient Boosting (XGBoost) \cite{Lyu2020, Wang2020}. While RF \cite{breiman2001random} uses bagging (\ie bootstrap aggregating) and random features subsets to train multiple and diverse trees independently, Gradient Boosting \cite{friedman2001greedy} builds one tree at a time, introducing a weak learner to improve shortcoming of existing trees by assigning more weight on instances with wrong predictions and high errors. XGBoost \cite{chen2016xgboost} is a specific and more efficient implementation of the GBT method. Other variants were also explored by Schroeder \etal \cite{Schroeder2014}, including Conditional trees, NB trees, and Functional trees.

Regression-based methods appeared in 11 selected papers, most of which adopted logistic regression \cite{Schroeder2014, Gnad2015, Soliman2015, Dong2016, Agajanian2018, Gumpinger2020, Lyu2020}. We also found papers using regularized regressions, including Ridge \cite{Nicora2019} and Lasso regression \cite{Davoli2013}. Regularization aims to discourage complex models by penalizing the magnitude of the coefficients and the error term. While Ridge regression forces variables with minor contributions to the model to have their coefficients close to zero, Lasso regression forces these coefficients to be exactly zero, thus keeping only the most significant variables in the final model. The works by Park \etal \cite{Park2015, Park2017} introduced new approaches based on linear regression and regularization schemes. Authors proposed a sparse overlapping group lasso to perform group selection while identifying crucial genes (\ie potential CDGs) within each group \cite{Park2015}, and %Their method extends the previous latent group lasso \cite{obozinski2011group}, which guarantees `groupwise sparsity' but includes in the model all genes in the selected group. By adding a sparse group lasso penalty, authors aim to also accomplish `within group sparsity', identifying fewer genes within each selected group that may act as driver genes.
%Their second paper \cite{Park2017} introduced 
an interaction-based feature-selection strategy with adaptive regularization that adjusts the amount of the L1-type penalty imposed on each gene proportionally to the degree to which gene expression alteration is explained by CNAs \cite{Park2017}.
%according to a score that measures how much of the gene expression alteration may be explained by CNAs%, such that genes with larger scores represent the most promising CDG candidates and receive smaller penalties.  

Probabilistic and artificial neural network (ANN) methods were less frequent among selected papers. The naïve Bayes algorithm was applied in two studies \cite{Schroeder2014, U2014}, while a Bayesian network was adopted only in one \cite{U2014}. U \etal \cite{U2014} used both the traditional implementation of naïve Bayes, as well as the DTNB algorithm that combines naïve Bayes with induction of decision tables \cite{hall2008combining}. Want \etal \cite{Wang2018} proposed a Bayesian hierarchical modeling approach to identify mutations that best predict alterations in gene expression, which represent candidate drivers. %was motivated by a previous Bayesian framework developed for performing eQTL mapping. Their method 
%aims to predict mRNA expression levels from somatic mutations and related functional genomic annotations, identifying the input variables (\ie mutations) that best predict the output (\ie expression), which represent candidate drivers for cancer. 
Furthermore, ANNs were used in three works \cite{U2014, Althubaiti2019, Wang2020}.
%U \etal \cite{U2014}, Althubaiti \etal \cite{Althubaiti2019} and Wang \etal \cite{Wang2020}. 
Althubaiti \etal \cite{Althubaiti2019} and Wang \etal \cite{Wang2020} trained an ANN with two hidden layers using a Rectified Linear Unit (ReLU) as an activation function for the hidden layers. In the former, the network receives as input embedding vectors generated from different ontologies, while in the latter, the input vectors are built from the pathogenicity prediction scores provided by \textit{in silico} tools.

Among the papers using supervised learning, it is worth highlighting three works \cite{Schroeder2014, U2014, Wang2020} that explored the greatest number of algorithms in their experiments. U \etal \cite{U2014} compared models trained by 11 distinct algorithms, including SVM-based, tree-based, neural network, and probabilistic methods. SVM presented the best performance among the classifiers, but a weighted voting approach among the 11 models achieved the most robust results. Schroeder \etal \cite{Schroeder2014} compared six algorithms, including regression-based, probabilistic, and tree-based, and observed that RF produced the highest predictive performance. Wang \etal \cite{Wang2020} analyzed seven algorithms, including SVM-based, tree-based, and ANN and found XGBoost as the top-performing method. %For the last class of ML methods, authors used four distinct algorithms.%U \etal \cite{U2014},
%and within the tree-based methods they used four different algorithms.

We also identified the use of less traditional supervised learning methods. Fu \etal \cite{Fu2012} used the Bayesian Factor Regression Modeling (BFRM), a sparse statistical model for high-dimensional data analysis \cite{wang2007bfrm}. U \etal \cite{U2014} also employed Decision Tables and Locally Weighted Learning (LWL), an instance-based algorithm that makes a prediction by taking a weighted neighborhood of the test instance and then building a classifier using this weighted subset.Finally, Han \etal \cite{Han2019} applied a ML principle, but the weight parameters in the proposed score are learned by maximizing their global weighted score stats from previous genes in the training mutation data.

%--
\subsection{\textbf{Methods based on deep learning}}

Despite the increasing use of deep learning in Bioinformatics, only four \cite{Tavanaei2017, Agajanian2019, Luo2019, Schulte-Sasse2019} papers adopted this class of algorithms. Convolutional neural networks (CNN) were used in three papers \cite{Tavanaei2017, Agajanian2019, Luo2019}. Tavanaei \etal \cite{Tavanaei2017} developed a parallel CNN with three branches followed by a multi-layer fully connected neural network. Each branch receives a projection of a 3-D protein structure to a 2-D feature map set %with dimensions of 200 $\times$ 200 
for 16 features associated with the atomic coordinates $< x, y, z >$. The input data is processed by four convolution and pooling layers adopting the ReLU activation function and three fully connected layers. %, including the final classifier. 
Luo \etal \cite{Luo2019} trained a 1-D CNN with a mutation-based feature matrix as input. %, which is constructed by considering the mutation profile of a gene and its neighbors in a similarity network. 
Authors varied the number of convolutional layers (1-4) and the number of fully connected layers (1-3), determining the best hyperparameters by grid search. %All convolutional layers adopted ReLU as the activation function.

Agajanian \etal \cite{Agajanian2019} trained a CNN model that processes encoded raw nucleotide sequences. The optimal CNN architecture was defined using a grid search over 72 different architectures. The authors evaluated label encoding (\ie each nucleotide receives a unique integer ID), word2vec embedding (\ie each nucleotide receives a numeric representation in a vector space based on the analysis of its sequential context), and one-hot encoding (\ie each nucleotide receives a bit encoded string), with the latter achieving best predictive performance. %According to their experiments, the one-hot encoded sequences yielded the best predictive performance. 
Their approach has the attractive property of combining DNA-derived scores computed by the CNN as an input feature to an RF model, along with other 32 features derived from FI predictors. Their results suggest that high-level features learned from genomic information using the CNN model complement the FI scores often employed in the field.%to classify cancer mutations.

Finally, Schulte-Sasse \etal \cite{Schulte-Sasse2019} used an extension of the CNN framework designed to handle graph-structured data as input, named Graph Convolutional Network (GCN) \cite{kipf2016gcn}. GCNs aim to directly classify the nodes of a network based on the characteristics of the nodes and the network structure. Authors adopted a PPI network and the feature vector associated with each node in the network contains information on gene mutation rates, gene expression, and DNA methylation collected for 16 cancer types. The optimal GCN found by the authors was composed of two graph convolutional layers with 50 and 100 filters, respectively.

%--
\subsection{\textbf{Methods based on unsupervised learning}}

Three unsupervised methods applied to CDGs prediction were identified in our survey. Since the collection of experimentally validated driver genes may be difficult and there is limited knowledge about it, these methods aim to identify candidate driver genes without relying on known examples of positive and negative drivers. Two of these works explored the concept of module discovery. Manolakos \etal \cite{Manolakos2014} proposed a robust and fast algorithm for detecting gene drivers using clustering to find genes with similar expression profiles across cancer patients. After gene clusters are created using the K-means algorithm, their method identifies sparse representations of genes in a given cluster as a linear combination of a small number of regulatory genes, pointed as potential cancer drivers. An Expectation-Maximization technique was used by Lu \etal \cite{Lu2018} in their module-based framework integrating transcriptome and genome. To identify CDGs, authors proposed an iterative approach that %learns a regulation program for each module, which involves 
determines the best modulators to explain gene expression profiles of genes in a given module and re-assigns each gene to the module whose associated
regulation program best predicts its behavior. In Xi \etal \cite{Xi2019}, a subspace learning framework was employed to obtain a low-dimensional vectorized representations of unannotated genes using a binary mutation matrix as input. %The input mutation data is provided as a binary matrix indicating the occurrence of mutations for each gene in all investigated samples. %, which is transformed into a low-dimensional subspace of gene representations by subspace learning. 
Driver genes can be discriminated evaluating the distances between the output vectors and the origin in the low-dimensional subspace, with the top-ranked genes according to their distance scores being promising driver candidates.

\subsection{\textbf{Methods vs. data categories}}

An analysis of Fig.~\ref{fig:algorithm-analysis}-b allows us to identify that while SVM-based, tree-based, and regression-based methods have been used in conjunction with all data categories defined in the current survey, other algorithms have found applications with specific types of features. ANNs have been applied solely over functional impact and ontology-based features, and were not used with data integration yet. Probabilistic methods and other less traditional supervised learning methods were used to train models from functional genomics, functional impact, and genomic variation data. Unsupervised learning was applied specifically to functional genomics, ontology-based, and genomic variation data. Finally, deep learning algorithms were applied to all data categories except for ontology-based.

  \begin{figure*}[!ht]
    \centering
    \includegraphics[width=\textwidth]{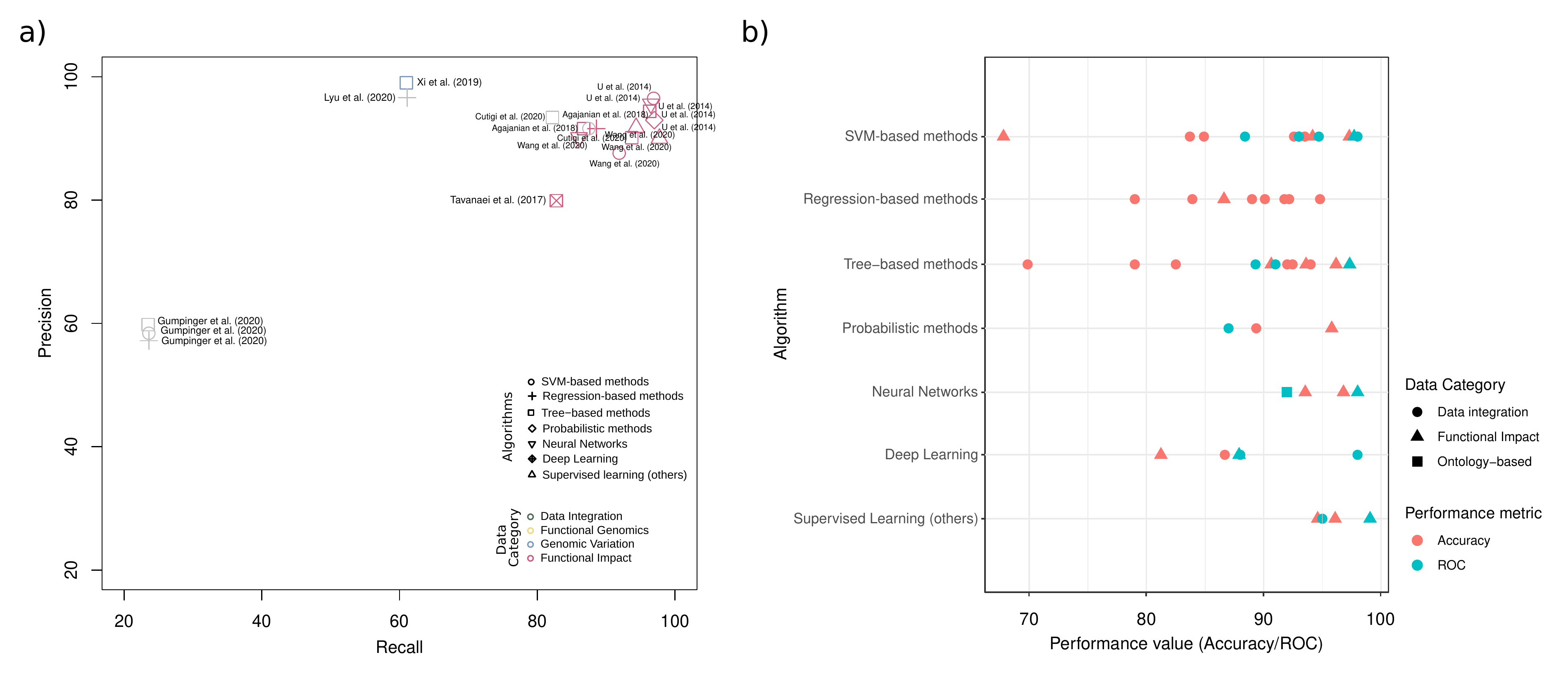}
    \caption{Summary of predictive performance reported in papers using supervised learning and deep learning, and its association with implementation details related to algorithm type and data category. a) Trade-off between precision and recall. b) Summary of accuracy and ROC values.}
    \label{fig:models_performance}
\end{figure*}

\subsection{\textbf{Model validation protocols}}

Among the methods based on supervised learning and deep learning, we observed the main model validation protocols. This is a crucial aspect of ML methodology, since a wrong validation may lead to over-optimistic expectations regarding the model's performance. Two papers \cite{Tavanaei2017, Celli2018} used the most basic validation protocol, which is the holdout method, \ie a simple division of the original dataset into train and test sets. %Tavanaei \etal \cite{Tavanaei2017} used a 85\%/15\% ratio, while Celli \etal \cite{Celli2018} adopted a 70\%/30\% ratio. 
A drawback of the holdout method is that the measured performance can highly depend on the instances included in the train and test datasets. Tavanaei \etal \cite{Tavanaei2017} repeated the holdout process three times using different random seeds to observe performance variation of their CNN model, nonetheless, standard deviations were not reported by authors. About 70.7\% of papers used the $k$-fold cross-validation process, which despite demanding  more computational power and time to run, it is the most recommended approach for model validation under limited data. Besides providing a performance distribution over $k$ evaluations, it ensures that every observation from the original dataset has the chance of appearing in the train and test set, providing a more robust performance assessment. Finally, some papers \cite{Manolakos2014, Agajanian2018, Agajanian2019, Luo2019, Nicora2019} combined the holdout and the $k$-fold cross-validation methods: the k-fold cross-validation is run over the train set derived from the holdout, and the best model built during this process is further applied over the test set. This is an interesting approach, as the performance obtained for the test set is more reliable since it is guarantee to be free from data leakage, which occurs when knowledge about the test set is shared with the train set during model development.

Additionally, some papers also evaluated their predictive models with independent test sets \cite{Tan2012, Mao2013, Schroeder2014, Anoosha2015, Soliman2015, Dong2016, Tavanaei2017, Zhou2018, Collier2019, Han2019, Luo2019, Colaprico2020, Lyu2020, Wang2020, Nulsen2021}. For instance, Mao \etal \cite{Mao2013} and Wang \etal \cite{Wang2020} generated four independent test sets to evaluate their approaches in and compare to existing methods. Collier \etal \cite{Collier2019} assessed the generalization properties of different methods using a subset of cancer genes from the Cancer Gene Census (CGC) v86 that were not included in the development of previous models. Han \etal \cite{Han2019} analyzed the ability of their model to recover a set of 99 high confidence cancer genes manually curated from literature. %Colaprico \etal \cite{Colaprico2020} performed analyses for several types of cancer, as well as for breast cancer molecular subtypes, to demonstrate the potential of their method. 
Nulsen \etal \cite{Nulsen2021} evaluated their method with an independent cancer cohort data of osteosarcoma, a rare bone cancer, and found that it was able to identify reliable cancer drivers in individual patients even for cancer types not used for training. Interestingly, we also found papers that carried out experimental analyses of selected findings to characterize mutations and their functional impact \cite{U2014, Wang2018, Han2019}.

\subsection{\textbf{Predictive performance of ML strategies}}

Among models built with supervised and deep learning methods, performance was assessed mainly using accuracy, the area under the ROC curve (ROC), recall (\ie sensitivity), and precision. Some papers also reported the area under the precision-recall curve (PRC) \cite{Schulte-Sasse2019, Lyu2020, Gumpinger2020} and the F1-score \cite{U2014, Soliman2015, Agajanian2018, Celli2018, Lu2018, Zhou2018, Agajanian2019, Althubaiti2019, Cutigi2020, Gumpinger2020, Wang2020}, which are metrics that summarize the trade-off between precision and recall. The PRC values varied from 41.65 \cite{Gumpinger2020} to 83.00 \cite{Schulte-Sasse2019}, while the F1-score ranged from 33.40 \cite{Gumpinger2020} to 99.00 \cite{Celli2018}. Moreover, three papers reported the Matthew's Correlation Coefficient (MCC) \cite{Soliman2015, Zhou2018, Schulte-Sasse2019} and one paper \cite{Tokheim2016} proposed the mean absolute log2 fold change (MLFC) metric, which quantifies the deviation between the theoretically expected p-values and the observed p-values generated by a method.%, with lower values indicating better statistical modeling of the passenger gene null distribution.

To investigate whether any association may exist between the model's predictive performance and implementation decisions regarding the type of algorithm or data used, we summarized the performance reported in the selected papers, either in the main text or supplementary materials. When multiple values were reported in a paper for the same combination of data category and algorithm type, we averaged them. Only accuracy, ROC, precision, and recall were observed for this analysis, which is summarized in  Fig~\ref{fig:models_performance}. We note that papers that adopted two or more data categories were classified as data integration in this analysis. In Fig~\ref{fig:models_performance}-a, we observe the trade-off between precision and recall. The highest recall values were obtained by papers using functional impact features, while the highest precision values are related to papers using data integration or genomic variation. Interestingly, the predictive performance does not seems to be segregated by algorithm type but rather by the features type. A good example for this observation is the work by U \etal \cite{U2014}, which used five distinct groups of algorithms to train models with the same feature vector and found a slight variation among their performances. 

Fig~\ref{fig:models_performance}-b summarizes accuracy and ROC values. We observe that except for regression-based methods and deep learning, the majority of the top-performing models were based on functional impact features. One approach using ANN and ontology-based features \cite{Althubaiti2019} achieved very competitive ROC values (\ie 91.96) despite relying on a single data category. Regarding the types of algorithms, we observed that SVM-based and Tree-based methods, the two most frequently used, had consistent performance, with most models surpassing the 90.00 mark. Moreover, ANN and other supervised learning techniques, although not as frequent as the previous approaches, also resulted in models with very high performance. Among these, we highlight models trained with Adaboost and ANN that achieved ROC values of 99.07 and 98.01, respectively \cite{Wang2020}.
%Finally, we emphasize that accuracy was the most used metric among selected papers, reported in 20 articles, followed by ROC score (13 papers).

\subsection{\textbf{Feature selection}}

In ML, feature selection (FS) reduces the input feature space by removing irrelevant, redundant, or noisy features. Dimensionality reduction aims to keep the features that contribute most to the prediction task, decreasing the chance of overfitting and leading to better performance. Moreover, FS also helps interpret the patterns learned by the model, understanding the relative usefulness of each feature towards predicting the target \cite{U2014}. Several selected papers adopted the concept of FS in their methodology \cite{Carter2009, Tan2012, Davoli2013, Mao2013, U2014, Anoosha2015, Gnad2015, Park2015, Soliman2015, Dong2016, Park2017, Tavanaei2017, Agajanian2018, Guan2018, Zhou2018, Lyu2020, Nulsen2021}. Among these, we observed the use of mutual information \cite{Carter2009}, DX score \cite{Tan2012}, Lasso regression \cite{Davoli2013}, Mann–Whitney U test \cite{Mao2013}, and Chi-square \cite{Soliman2015}.

U \etal \cite{U2014} applied five FS algorithms (\ie oneR, reliefF, Chi-square, gain ratio, and correlation-based) and retained all the top features indicated by at least three algorithms. Mao \etal \cite{Mao2013} evaluated all possible combinations with fewer than four features using k-fold cross-validation and ROC score, and the best feature subset was then expanded using a hill-climbing strategy to iteratively include the remaining features into the combination. Agajanian \etal \cite{Agajanian2018} applied a recursive feature elimination process, in which features are removed one by one and the model is trained on the resulting data set; if the accuracy remains above a predefined threshold, the feature is removed permanently, and the process is repeated. A similar process was used in Zhou \etal \cite{Zhou2018} and all feature's removal that resulted in a better MCC value with a cutoff of 0.5 was permanently discarded. Guan \etal \cite{Guan2018} proposed an FS procedure that leverages the correlation between the predictors and the gene essentiality on two complementary scales, global and local, and keeps the nine most predictive gene expression features and the top predictive CNA feature for each gene. Anoosha \etal \cite{Anoosha2015} adopted a ranker search method based on features evaluation by an SVM, which ranks features by the square of the weight assigned by the model.

Some papers analyzed feature's importance separately for oncogenes and TSGs \cite{Davoli2013, Gnad2015, Lyu2020}. Davoli \etal \cite{Davoli2013} %adopted Lasso regression, which performs embedded FS, 
identified the most predictive features for oncogenes and TSGs by Lasso regression, % and further compared their distributions among oncogenes, TSGs, and neutral genes in a pairwise manner. When comparing oncogenes and TSGs, the authors identified 
identifying significantly higher levels of amplification frequency and significantly lower levers of deletion frequency, loss-of-function(LOF)/benign mutation ratio, and splicing/benign mutation ratio in oncogenes. Moreover, the LOF/benign ratio and the missense entropy were the best features to discriminate among oncogenes and TSGs in their model. Gnad \etal \cite{Gnad2015} conducted a similar analysis, finding that the most informative predictors for oncogenes were the occurrence of mutation hotspots, the HiFI/LoFI missense mutations ratio, and the amplification frequency, whereas for TSG prediction, the LOF/benign ratio, splicing/benign ratio, and the frequency of homozygous copy number losses contributed most. Lyu \etal \cite{Lyu2020} grouped correlated features using hierarchical clustering prior to FS, identifying three and five relevant feature groups for TSGs and OGs.

\section{Challenges and Perspectives}

Undoubtedly, the use of ML algorithms to predict cancer driver genes and mutations has enabled important scientific advances and has an excellent potential to go even further. Our survey has shed light on the broad data and algorithmic landscape behind this problem, as well as the potentials and gaps in the myriad possibilities of combining them into novel solutions. Nonetheless, to accelerate the computational discovery of new cancer drivers, several challenges remain to be explored. % in future works. %and represent interesting future research directions.

\subsection{\textbf{Class imbalance}}

The prediction of CDGs and driver mutations is an inherent class imbalance problem due to the low number of known drivers compared to the passenger ones. Most supervised algorithms work more effectively for balanced train sets. Thus, highly imbalanced datasets are a long-standing challenge to ML since models tend to be biased towards the majority class and produce large error rates for the minority class \cite{leevy2018survey}. Despite the important impact on the model development process, few works using supervised or deep learning approaches have discussed strategies to deal with this issue. Among the selected works, we observed the use of downsampling of the majority class \cite{Capriotti2011, Tan2012, Gnad2015, Luo2019, Cutigi2020, Gumpinger2020}, weighted SVM \cite{Mao2013}, cost-sensitive classifier \cite{U2014}, and weighted cross-entropy as loss function \cite{Schulte-Sasse2019}. One-class SVM, although not originally designed to deal with class imbalance, may be particularly useful for highly imbalanced datasets \cite{Collier2019, Nulsen2021}.

Another related issue is the choice of appropriate performance metrics. We observed a prominent use of accuracy and ROC scores for performance assessment, which under skewed class distributions do not provide adequate assessment for the minority class (\ie driver genes or mutations). Both metrics are less sensitive to false positives as the size of the negative class grows and can produce misleadingly high values. %Several works adopted these metrics without adequately dealing with class imbalance, resulting in unrealistic predictive performance. 
Thus, using metrics that can pay proper attention to methods' performance for the minority class is crucial, such as precision, recall, their tradeoff summarized by F1 score and PRC, and MCC. We emphasize that this methodological aspect of utmost relevance in the model development process cannot be neglected in the development of future models. 

% false positive rate used by ROC, FPR=FPFP+TN, will be less sensitive to changes in FP as the size of the negative class grows.

\subsection{\textbf{Data leakage}}

Data leakage happens when information from the test set is accidentally used to develop the model, which may happen in subtle ways. For instance, pre-processing the complete dataset with normalization, standardization, or data imputation techniques before data partitioning (\eg holdout or k-fold cross-validation) cause data points from test set to influence in the train set. Similarly, performing feature selection in the complete dataset that will further be divided into independent partitions for model training and validation produces the same unintentional effect. Even under the use of cross-validation, data leakage may occur if the same test fold used to optimize the model with hyperparameter tuning or feature selection, for instance, is also applied for model evaluation and selection. These methodological flaws were identified in several revised papers. Taking steps to prevent data leakage is imperative to guarantee better generalization of models. Data preparation pipelines should be modeled based on training data only and then applied to all partitions (\eg train, test, and validation set). %Data pre-processing should be restricted for training data 
Nested cross-validation is a promising alternative for comparing and selecting ML models with size-limited datasets, reducing the bias in the error estimate when both hyperparameter tuning and evaluation are carried out \cite{varma2006bias, vabalas2019machine}. Moreover, FS protocols for high-dimensional data have been proposed to minimize optimistic performance estimates \cite{kuncheva2018feature}, especially under low-sample-size as it is the case for some cancer types.

\subsection{\textbf{Graph-based machine learning}}

Networks, or graphs, are pervasive in biology, representing the existing interactions between genes, gene products, and other molecules. Network-based analysis has attracted considerable attention in the prediction of CDGs. Some of the selected papers adopted node embedding and network propagation algorithms. Others used graph theoretic techniques to compute the network's node properties (\eg degree, betweenness) used as inputs for ML algorithms. There is also a large body of works using network analysis without interface with ML, which are out of our scope. Thus, a natural and still underexplored direction in this domain is the use of graph-based ML algorithms, which can directly process graph-structured data in an end-to-end manner without the need for handcrafted feature engineering. 
Graph Neural Networks (GNNs) \cite{wu2020comprehensive} can better explore network's topological information for node-level or edge-level prediction tasks in contrast to traditional network analysis, finding several fruitful applications in Bioinformatics \cite{zhang2021graph}. However, only one selected paper adopted this approach.  Schulte-Sasse \etal \cite{schulte2021integration} recently extended their previous work included in our survey \cite{Schulte-Sasse2019}, integrating multi-omics data and PPI networks into a learning framework based on Graph Convolutional Networks (GCNs), a variant of GNNs. Their model predicted CDGs with average PRC values higher than tools exploring network-based analysis, ML-based classification of omics data, or a combination of both. We believe that further exploration of GNNs with multimodal data, proposing strategies to improve robustness to structural noise and class imbalance, may enable an even more comprehensive investigation and precise prediction of CDGs.

\subsection{\textbf{Data representativeness}}

In ML, guaranteeing that all the variability associated with a prediction task is represented in the training dataset is as essential as data volume. In the context of cancer, the high intra- and inter-tumor heterogeneity pose additional challenges to data representativeness. Data may not be evenly distributed across distinct cancer types, tumor stages, patients' clinical profiles, or even distinct demographic groups (\eg gender and ethnicity), which certainly interferes with model generalization power. For instance, sex-associated differences in tumor molecular profiles and mutation frequency were previously reported \cite{li2018cr}. Likewise, mutational processes vary widely among cancer types \cite{brown2019finding}. Thus, any underrepresented subpopulation may suffer from biased prediction results when building models without considering this inequality. 

This rationale applies not only to a sample-oriented perspective but also to a gene-oriented perspective. Here, we highlight two possible underrepresentation issues. First, some driver genes are commonly mutated across cancer types, while others are tumor-specific \cite{poulos2019finding}. Thus, tumor-specific drivers may not be identified in pan-cancer models due to low statistical power arising from low-sample size.  Patterns from local samples tend to be diluted among patterns from the population in consideration, hindering their identification by the ML algorithm.  %This is specially true for more rare cancers. 
Second, known cancer driver mutations tend to occur in a subset of human genes, while negative samples are spread across the genome. As discussed by Raimondi \etal \cite{raimondi2021current}, this may cause ML models to actually learn the simpler gene-level patterns instead of the more intricate molecular-level functional effects of driver variants, achieving an unrealistic good performance for variant-level prediction. Thus, we stress the importance of defining high-quality, unbiased negative class samples for supervised learning methods. Finally, for further advancement of the field, new computational strategies are needed to overcome the underrepresentation issues. We point as promising directions training patient-specific, cohort-specific, or cancer-specific models.

\subsection{\textbf{Model interpretability}}

Although some works have investigated the most relevant features to predict CDGs, and more specifically, TSGs and OGs, model interpretability is still in its infancy within this domain. Model interpretability helps increase the predictive model's trust and provides valuable biological insights about molecular differences among driver and passenger mutations that may fill current knowledge gaps. Model interpretation is especially challenging with deep learning and graph-based learning, but several recent works have proposed strategies to comprehend the decision-making process itself, exploring their applications to bioinformatics \cite{schulte2021integration, talukder2021interpretation}. We advocate the use of model interpretability tools in future works as they may explain the discovered patterns and, consequently, ensure these patterns are significant and consistent with the target task (\ie avoid \textit{Clever Hans} effect \cite{lapuschkin2019unmasking}), elucidate properties associated to subgroups of drivers, and point to particular cases that deserve further attention from experts. However, we raise awareness that pitfalls in model interpretability may produce incorrect conclusions \cite{molnar2020general} just like any other step in ML methodology, thus demanding careful use.

%result in faster adoptability
\subsection{\textbf{Non-coding drivers}}
Mutations occurring in both coding and non-coding DNA regions may play a crucial role in cancer development. For instance, highly recurrent mutations in the promoter region of \textit{TERT} gene have been found in more than 50 tumor types, prompting efforts to identify additional non-coding driver events \cite{elliott2021non, bell2016understanding}. However, there are at present few computational tools specifically tailored for detecting drivers in non-coding regions (\eg \cite{funseq2, lawrence2013mutational, guo2020mutspot}). Identifying signals of positive selection in non-coding DNA is even more challenging because the non-coding region is about 50 times larger than the coding exome, and the number of known cancer genes with non-coding mutations is much more limited \cite{belkadi2015whole, elliott2021non}. In contrast, the more frequent use of whole-genome sequencing (WGS) tends to produce more and more data that could be analyzed for this purpose. Given the several ways in which variations in non-coding DNA may contribute to tumor emergence, exploring ML learning (especially unsupervised algorithms) and the increasing volume of WGS data represents an important research opportunity to advance in the field.

\section{Conclusion}
Years of research towards understanding the patterns associated with cancer driver events and proposing strategies to distinguish them from somatic changes have resulted in several crucial advances in identifying genetic factors related to the emergence and development of tumors. However, this field is far from being completely understood, with many biological and technical challenges still limiting our capacity to detect cancer drivers comprehensively. For the research field to remain advancing, it is imperative to look back, map out the significant advances and the methodological details entailed in previous works, and outline fruitful perspectives based on remaining gaps. Our survey aimed to summarize efforts related to ML-based methods, which have been the engine behind many successful computational approaches to predict cancer drivers. We hope that the panoramic and integrated view on genomic datasets and ML algorithms provided by our work can help advance this research field, directing future work towards challenges that remain open to improve our ability to identify mutations and genes driving carcinogenesis.

\section{Competing interests}
There is NO Competing Interest.

%\section{Author contributions statement}
%Must include all authors, identified by initials, for example:
%S.R. and D.A. conceived the experiment(s),  S.R. conducted the experiment(s), S.R. and D.A. analysed the results.  S.R. and D.A. wrote and reviewed the manuscript.

\section{Acknowledgments}
%The authors thank the anonymous reviewers for their valuable suggestions. 
This work has been supported in part by funding from Conselho Nacional de Desenvolvimento Científico e Tecnológico (CNPq) and Coordena\c{c}\~{a}o de Aperfei\c{c}oamento de Pessoal de N\'{i}vel Superior (CAPES) - Finance Code 001.

\bibliographystyle{unsrt}
\bibliography{references}

%USE THE BELOW OPTIONS IN CASE YOU NEED AUTHOR YEAR FORMAT.
%\bibliographystyle{abbrvnat}
%\bibliography{reference}

%% sample for biography with author's image
% \begin{biography}{{\color{black!20}\rule{77pt}{77pt}}}{\author{Renan Andrades.} Graduated in Electronic Engineer from the Universidade Federal de Pelotas (UFPel) and master degree student at Universidade Federal do Rio Grande do Sul (UFRGS). Researcher with the Bioinformatics Core at Hospital de Cl\'inicas de Porto Alegre, Porto Alegre, Brazil. His research interests include Machine Learning, Data Science and Bioinformatics.}
% \end{biography}
\begin{biography}{}{\author{Renan Andrades.} Master's student in the Post-Graduate Program in Computer Science (PPGC) from the Universidade Federal do Rio Grande do Sul (UFRGS) and Researcher with the Bioinformatics Core at Hospital de Cl\'inicas de Porto Alegre, Porto Alegre, Brazil. He received his B.S. degree in Electronic Engineer from the Universidade Federal de Pelotas (UFPel). His research interests include Machine Learning, Data Science, and Bioinformatics.}
\end{biography}

%% sample for biography without author's image
\begin{biography}{}{\author{Mariana Recamonde-Mendoza.} Associate Professor with the Institute of Informatics at Universidade Federal do Rio Grande do Sul (UFRGS) and Researcher with the Bioinformatics Core at Hospital de Cl\'inicas de Porto Alegre, Porto Alegre, Brazil. She received her PhD degree in Computer Science from UFRGS in 2014.  Her research interests include Bioinformatics, Computational Biology, and Machine Learning. }
\end{biography}

\end{document}